\documentclass[runningheads]{llncs}
\usepackage{graphicx}
\usepackage{comment}
\usepackage{amsmath,amssymb} % define this before the line numbering.
\usepackage{color}

\usepackage{bm}
\usepackage{multirow}
\usepackage{booktabs}
\usepackage{float}
\usepackage{url}
\usepackage[colorlinks,linkcolor=blue]{hyperref}
\begin{document}
% \linenumbers
\pagestyle{headings}
\mainmatter

\title{Progressive Point Cloud Deconvolution Generation Network} % Replace with your title

%\begin{comment}
%\titlerunning{ECCV-20 submission ID \ECCVSubNumber} 
%\authorrunning{ECCV-20 submission ID \ECCVSubNumber} 
%\author{Anonymous ECCV submission}
%\institute{Paper ID \ECCVSubNumber}
%\end{comment}
%******************

\titlerunning{Progressive Point Cloud Deconvolution Generation Network}

%\author{First Author\inst{1}\orcidID{0000-1111-2222-3333} 
\author{Le Hui \and Rui Xu \and Jin Xie \and Jianjun Qian \and Jian Yang}
\authorrunning{L.Hui, X.Rui, J.Xie, J.Qian, J.Yang}
\institute{
	Key Lab of Intelligent Perception and Systems for High-Dimensional Information of Ministry of Education \\ Jiangsu Key Lab of Image and Video Understanding for Social Security
	\\ PCA Lab, School of Computer Science and Engineering \\ Nanjing University of Science and Technology
	\\ \url{https://github.com/fpthink/PDGN}
	\\ \email{\{le.hui, xu\_ray, csjxie, csjqian, csjyang\}@njust.edu.cn}}
%\end{comment}
%******************
%******************
\maketitle

\begin{abstract}
In this paper, we propose an effective point cloud generation method, which can generate multi-resolution point clouds of the same shape from a latent vector. Specifically, we develop a novel progressive deconvolution network with the learning-based bilateral interpolation. The learning-based bilateral interpolation is performed in the spatial and feature spaces of point clouds so that local geometric structure information of point clouds can be exploited. Starting from the low-resolution point clouds, with the bilateral interpolation and max-pooling operations, the deconvolution network can progressively output high-resolution local and global feature maps. By concatenating different resolutions of local and global feature maps, we employ the multi-layer perceptron as the generation network to generate multi-resolution point clouds. In order to keep the shapes of different resolutions of point clouds consistent, we propose a shape-preserving adversarial loss to train the point cloud deconvolution generation network. Experimental results demonstrate the effectiveness of our proposed method.
\keywords{Point cloud generation, GAN, deep learning, deconvolution network}
\end{abstract}

%-------------------------------------------------------
\section{Introduction}\label{introduction}
With the development of 3D sensors such as LiDAR and Kinect, 3D geometric data are widely used in various kinds of computer vision tasks. Due to the great success of generative adversarial network (GAN) \cite{goodfellow2014generative} in the 2D image domain, 3D data generation \cite{wang2018pixel2mesh,choy20163d,fan2017point,kulkarni20193d,tulsiani2018factoring,groueix2018papier,zamorski2018adversarial,yang2018foldingnet,zhao20193d} has been receiving more and more attention. Point clouds, as an important 3D data type, can compactly and flexibly characterize geometric structures of 3D models. Different from 2D image data, point clouds are unordered and irregular. 2D generative models cannot be directly extended to point clouds. Therefore, how to generate realistic point clouds in an unsupervised way is still a challenging and open problem.

Recent research efforts have been dedicated to 3D model generation. Based on the voxel representation of 3D models, 3D convolutional neural networks (3D CNNs) can be applied to form 3D GAN \cite{wu2016learning} for 3D model generation. Nonetheless, since the 3D CNNs on the voxel representation requires heavy computational and memory burdens, the 3D GAN is limited to generate low-resolution 3D models. Different from the regular voxel representation, point clouds are spatially irregular. Therefore, CNNs cannot be directly applied on point clouds to form 3D generative models. Inspired by PointNet \cite{qi2017pointnet} that can learn compact representation of point clouds, Achlioptas \emph{et al.}~\cite{achlioptas2017learning} proposed an auto-encoder based point cloud generation network in a supervised manner. Nonetheless, the generation model is not an end-to-end learning framework.
Yang \emph{et al.}~\cite{pointflow} proposed the PointFlow generation model, which can learn a two-level hierarchical distribution with a continuous normalized flow. Based on graph convolution, Valsesia \emph{et al.}~\cite{valsesia2018learning} proposed a localized point cloud generation model. Dong \emph{et al.}~\cite{shu20193dpc} developed a tree structured graph convolution network for point cloud generation. Due to the high computational complexity of the graph convolution operation, training the graph convolution based generation models is very time-consuming.

In this paper, we propose a simple yet efficient end-to-end generation model for point clouds. We develop a progressive deconvolution network to map the latent vector to the high-dimensional feature space. In the deconvolution network, the learning-based bilateral interpolation is adopted to enlarge the feature map, where the weights are learned from the spatial and feature spaces of point clouds simultaneously. It is desirable that the bilateral interpolation can capture the local geometric structures of point clouds well with the increase of the resolution of generated point clouds. % With the bilateral interpolation and max-pooling operations followed by the multi-layer perceptron (MLP) on the low-resolution point clouds, the local and global feature maps of the high-resolution point clouds are obtained, respectively. The concatenation of the local and global feature maps is used as the output of the deconvolution network.
Following the deconvolution network, we employ the multi-layer perceptron (MLP) to generate spatial coordinates of point clouds. By stacking multiple deconvolution networks with different resolutions of point clouds as the inputs, we can form a progressive deconvolution generation network to generate multi-resolution point clouds. Since the shapes of multi-resolution point clouds generated from the same latent vector should be consistent, we formulate a shape-preserving adversarial loss to train the point cloud deconvolution generation network. Extensive experiments are conducted on the ShapeNet~\cite{chang2015shapenet} and ModelNet~\cite{wu20153d} datasets to demonstrate the effectiveness of our proposed method.

The main contributions of our work are summarized as follows:
\begin{itemize}
	\item We present a novel progressive point cloud generation framework in an end-to-end manner.
	\item We develop a new deconvolution network with the learning-based bilateral interpolation to generate high-resolution feature maps.
	\item We formulate a shape-preserving loss to train the progressive point cloud network so that the shapes of generated multi-resolution point clouds from the same latent vector are consistent.
\end{itemize}
The rest of the paper is organized as follows: Section 2 introduces related work. In Section 3, we present the progressive end-to-end point cloud generation model. Section 4 presents experimental results and Section 5 concludes the paper.

%-------------------------------------------------------
\section{Related Work}\label{realted_work}

\subsection{Deep Learning on 3D Data}
3D data can be represented by multi-view projections, voxelization and  point clouds. Based on these representations, existing 3D deep learning methods  can be mainly divided into two classes. One class of 3D deep learning methods ~\cite{su2015multi,wu20153d,maturana2015voxnet,qi2016volumetric} convert the geometric data to the regular-structured data (i.e., 2D image and 3D voxel) and apply existing deep learning algorithms to them. The other class of methods ~\cite{monti2017geometric,simonovsky2017dynamic,landrieu2018large,te2018rgcnn,qi2017pointnet,qi2017pointnet++} mainly focus on constructing special operations that are suitable to the unstructured geometric data for 3D deep learning.

In the first class of 3D deep learning methods, view-based methods represent the 3D object as a collection of 2D views so that the standard CNN can be directly applied. Specifically, the max-pooling operation across views is used to obtain a compact 3D object descriptor~\cite{su2015multi}.  Voxelization~\cite{wu20153d,maturana2015voxnet} is another way to represent the 3D geometric data with regular 3D grids. Based on the voxelization representation, the standard 3D convolution can be easily used to form the 3D CNNs. Nonetheless, the voxelization representation usually leads to the heavy burden of memory and high computational complexity because of the computation of the 3D convolution. In addition, Qi \emph{et al.}~\cite{qi2016volumetric} proposed to combine the view-based and voxelization-based deep learning methods for 3D shape classification.

In 3D deep learning, variants of deep neural networks are also developed to characterize the geometric structures of 3D point clouds.  \cite{simonovsky2017dynamic,te2018rgcnn} formulated the unstructured point clouds as the graph-structured data and employed the graph convolution to form the 3D deep learning representation. %Landrieu \emph{et al.}~\cite{landrieu2018large} proposed superpoint graph for large-scale point clouds of millions of points.
Qi \textit{et al}.~\cite{qi2017pointnet} proposed PointNet that treats each point individually and aggregates point features through several MLPs followed by the max-pooling operation. Since PointNet cannot capture the local geometric structures of point clouds well, Qi \emph{et al.}~\cite{qi2017pointnet++} proposed PointNet++ to learn the hierarchical feature representation of point clouds.  By constructing the $k$-nearest neighbor graph, Wang \emph{et al.}~\cite{wang2018dynamic} proposed an edge convolution operation to form the dynamic graph CNN for point clouds. Li \textit{et al.}~\cite{li2018pointcnn} proposed PointCNN for feature learning from point clouds, where an $\chi$-transform is learned to form the $\chi$-convolution operation. 

\subsection{3D Point Cloud Generation}
%In recent years, deep generative models have achieved remarkable success. There are two powerful deep generative models: variational autoencoder (VAE)~\cite{kingma2013auto} and generative adversarial networks (GANs)~\cite{goodfellow2014generative}. There are many advanced methods~\cite{Sun2018PointGrowAL,sharma2016vconv,girdhar2016learning,chen2003visual,kazhdan2003rotation} that consider generating point clouds.

Variational auto-encoder (VAE) is an important type of generative model. Recently, VAE has been applied to point cloud generation. Gadelha~\emph{et al.}~\cite{gadelha2018multiresolution} proposed MRTNet (multi-resolution tree network) to generate point clouds from a single image. Specifically, using a VAE framework, a 1D ordered list of points is fed to the multi-resolution encoder and decoder to perform point cloud generation in unsupervised learning. 
 Zamorski~\emph{et al.}~\cite{zamorski2018adversarial} applied the VAE and adversarial auto-encoder (AAE) to point cloud generation. Since the VAE model requires the particular prior distribution to make KL divergence tractable, the AAE is introduced to learn the prior distribution by utilizing adversarial training. Lately, Yang \emph{et al.}~\cite{pointflow} proposed a probabilistic framework (PointFlow) to generate point clouds by modeling them as a two-level hierarchical distribution. As mentioned in PointFlow~\cite{pointflow}, it converges slowly and fails for the cases with many thin structures (like chairs).

Generative adversarial network (GAN) has achieved great success in the field of image generation~\cite{arjovsky2017wasserstein,denton2015deep,mao2017least,radford2015unsupervised,Mirza2014Conditional}. Recently, a series of attractive works~\cite{fan2017point,choy20163d,gwak2017weakly,yang20173d,shu20193dpc} ignite a renewed interest in the 3D object generation task by adopting CNNs. Wu~\emph{et al.}~\cite{wu2016learning} first proposed 3D-GAN, which can generate 3D objects from a probabilistic space by using the volumetric convolutional network and GAN. However, due to the sparsely occupied 3D grids of the 3D object, the volumetric representation approach usually faces a heavy memory burden, resulting in the high computational complexity of the volumetric convolutional network. To alleviate the memory burden, Achlioptas \emph{et al.}~\cite{achlioptas2017learning} proposed a two-stage deep generative model with an auto-encoder for point clouds. It first maps a data point into its latent representation and then trains a minimal GAN in the learned latent space to generate point clouds. 
However, the two-stage point cloud generation model cannot be trained in the end-to-end manner. Based on graph convolution, Valsesia \emph{et al.}~\cite{valsesia2018learning} focused on designing a graph-based generator that can learn the localized features of point clouds.  Similarly, Shu \emph{et al.}~\cite{shu20193dpc} developed a tree structured graph convolution network for 3D point cloud generation. The graph convolution based point cloud generation model can obtain the impressive results.

\begin{figure}
	\begin{center}
		\includegraphics[width=0.9\linewidth]{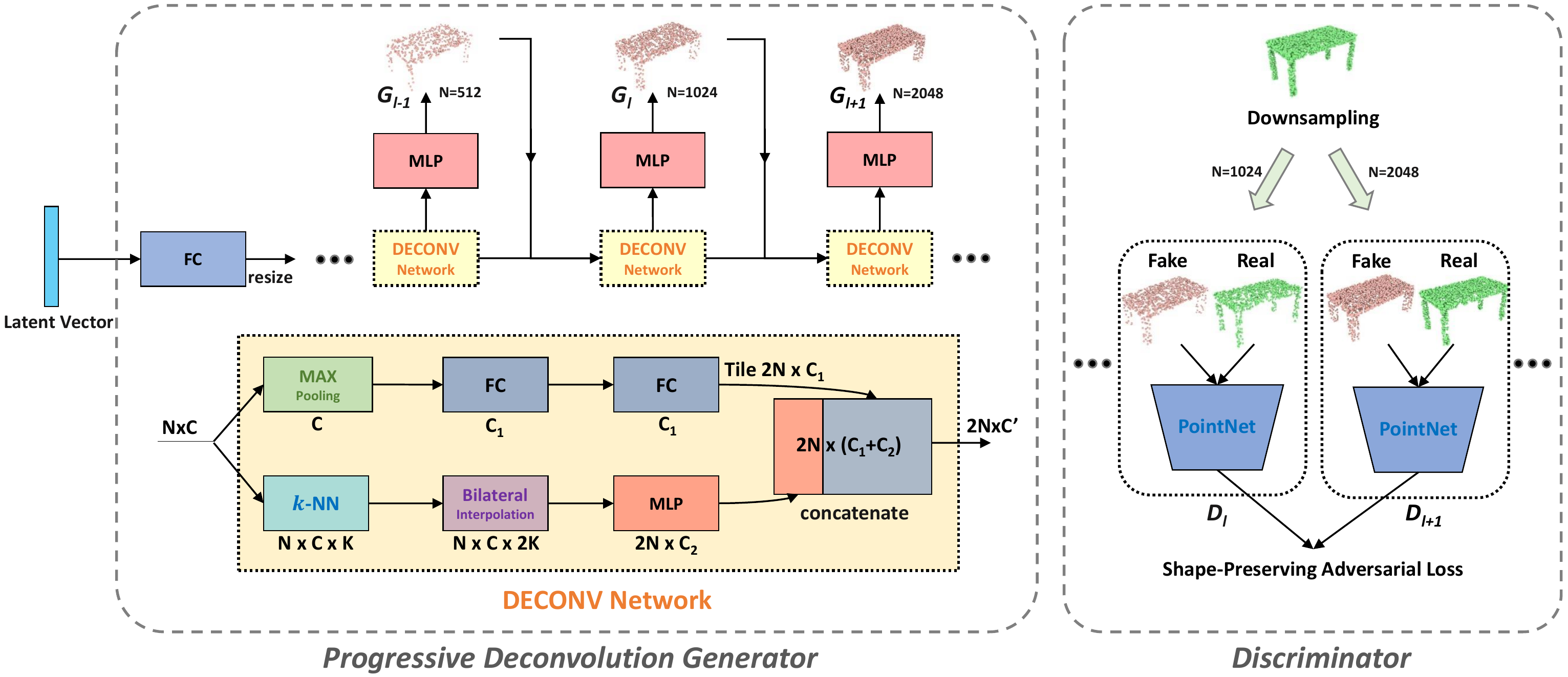}
	\end{center}
	\caption{The architecture of our progressive point cloud framework. The progressive deconvolution generator aims to generate point clouds, while the discriminator distinguishes it from the real point clouds. In the generator, the FC layer first maps the latent vector to a low-dimensional feature space. After resizing, the deconvolution network (``DECONV Network'') then progressively enlarges the feature map to the high-dimensional space, in which a learning-based bilateral interpolation is formed in the spatial and feature spaces. Following the deconvolution network, we apply a multi-layer perception (MLP) to generate 3D point clouds. In the discriminator, a shape-preserving adversarial loss is applied to the two adjacent resolutions to keep their shapes consistent.%Note that our architecture can be easily implemented in modern deep learning platforms.
	}
	\label{fig:network_generator}
\end{figure}

%-------------------------------------------------------
\section{Our Approach}\label{our_approach}
In this section, we present our progressive generation model for 3D point clouds. The framework of our proposed generation model is illustrated in Fig.~\ref{fig:network_generator}. In Section 3.1, we describe how to construct the proposed progressive deconvolution generation network. In Section 3.2, we present the details of the shape-preserving adversarial loss to train the progressive deconvolution generation network.

\subsection{Progressive deconvolution generation network}
Given a latent vector, our goal is to generate high-quality 3D point clouds. One key problem in point cloud generation is how to utilize a one-dimensional vector to generate a set of 3D points consistent with the 3D object in geometry. To this end, we develop a special deconvolution network for 3D point clouds, where we first obtain the high-resolution feature map with the learning-based bilateral interpolation and then apply MLPs to generate the local and global feature maps. It is desirable that the fusion of the generated local and global feature maps can characterize the geometric structures of point clouds in the high-dimensional feature space.

\begin{figure}[t]
	\begin{center}
		\includegraphics[width=0.85\linewidth]{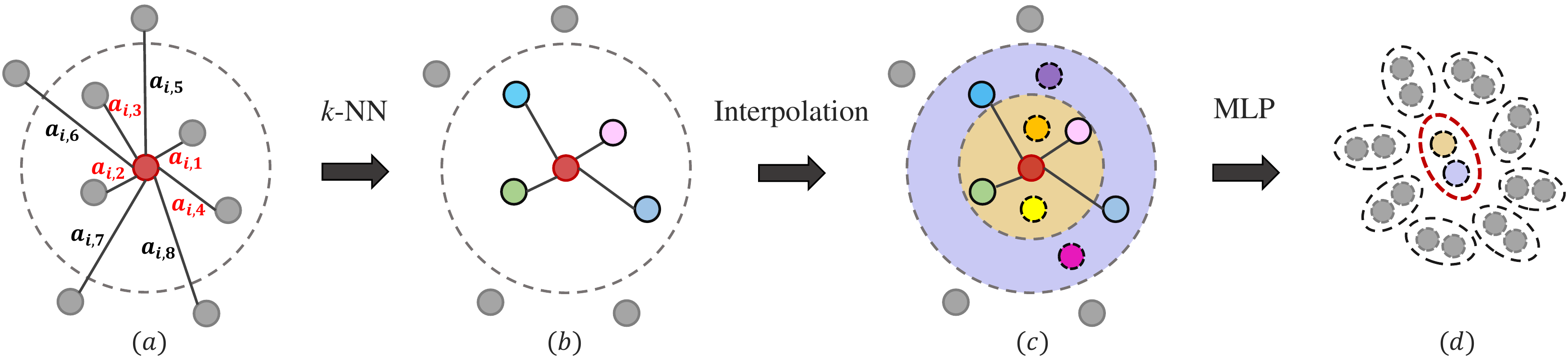}
	\end{center}
	\caption{The procedure of deconvolution network. %to create a new magnified feature map to generate 3D point clouds.
		First, we define the similarity between point pairs in the feature space~(a). We choose the $k$ nearest neighbor points ($k$-NN) in the feature space with the defined similarity in~(b). Then we interpolate in the neighborhood to form an enlarged feature map in~(c). Finally, we apply the MLP to generate new high-dimensional feature maps in (d). Note that we can obtain double numbers of points through the deconvolution network.}
	\label{fig:deconv}
\end{figure}

%\subsubsection{Construct Neighborhood}
\textbf{Learning-based bilateral interpolation.} Due to the disordered and irregular structure of point clouds, we cannot directly perform the interpolation operation on the feature map. Therefore, we need to build a neighborhood for each point on the feature map to implement the interpolation operation. In this work, we simply employ the $k$-nearest neighbor ($k$-NN) to construct the neighborhood of each point in the feature space. Specifically, given an input with $N$ feature vectors $\bm{x}_{i} \in\mathbb{R}^{d}$, the similarity between points $i$ and $j$ is defined as:
\begin{equation}
a_{i,j} = \exp\left(-\beta \left\|\bm{x}_{i} - \bm{x}_{j}\right\|_{2}^{2} \right)
\label{equ:edge_weight}
\end{equation}
where $\beta$ is empirically set as $\beta=1$ in our experiments. As shown in Fig.~\ref{fig:deconv}~(a) and (b), we can choose $k$ nearest neighbor points in the feature space with the defined similarity. And the parameter $k$ is set as $k=20$ in this paper.

%\subsubsection{Bilateral Interpolation}
Once we obtain the neighborhood of each point, we can perform the interpolation in it. As shown in Fig.~\ref{fig:deconv}~(c), with the interpolation, $k$ points in the neighborhood can be generated to $2k$ points in the feature space. Classical interpolation methods such as linear and bilinear interpolations are non-learning interpolation methods, which cannot be adaptive to different classes of 3D models during the point cloud generation process. Moreover, the classical interpolation methods does not exploit neighborhood information of each point in the spatial and feature space simultaneously.

\begin{figure}[t]
	\begin{center}
		\includegraphics[width=0.5\linewidth]{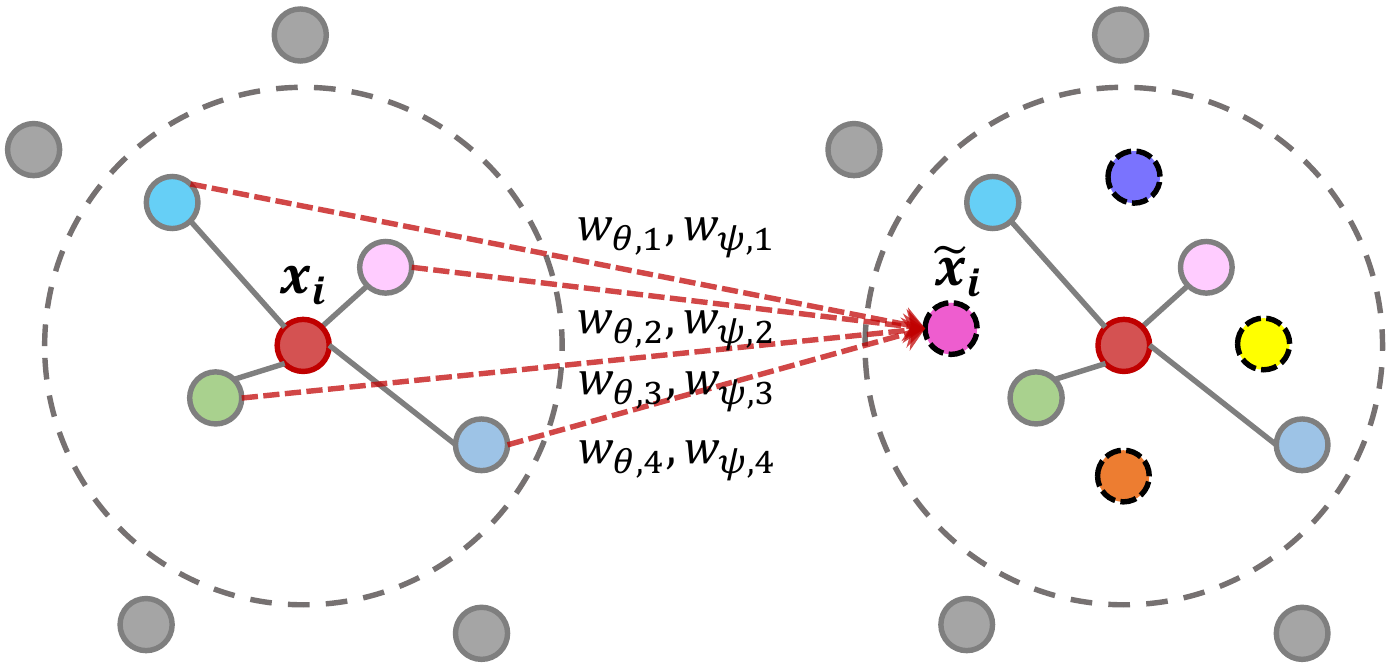}
	\end{center}
	\caption{The diagram of the learning-based bilateral interpolation method. The points in the neighborhood of the center point $\bm{x}_i$ are colored. We interpolate new points by considering the local geometric features of the points in the neighborhood. The $\bm{W}_{\theta,j}$ and $\bm{W}_{\psi,j}$, $j=1,2,3,4$ are the weights in the spatial and feature spaces to be learned.%, respectively
	}
	\label{fig:interpolation}
\end{figure}

To this end, we propose a learning-based bilateral interpolation method that utilizes the spatial coordinates and feature of the neighborhood of each point to generate the high-resolution feature map. %Specifically, as illustrated in Fig.~\ref{fig:interpolation},
Given the point $\bm{p}_i \in \mathbb{R}^{3}$ and $k$ points in its neighborhood, we can formulate the bilateral interpolation as:
\begin{equation}
\bm{\tilde{x}}_{i, l} =  \frac{\sum\nolimits_{j=1}^{k}\theta_{l}\left(\bm{p}_i, \bm{p}_j\right)\psi_{l}\left(\bm{x}_i, \bm{x}_j\right)\bm{x}_{j,l}}{\sum\nolimits_{j=1}^{k} \theta_{l}\left(\bm{p}_i, \bm{p}_j\right)\psi_{l}\left(\bm{x}_i, \bm{x}_j\right)}
\label{equ:bilateral}
\end{equation}
where $\bm{p}_i$ and $\bm{p}_j$ are the 3D spatial coordinates, $\bm{x}_i$ and $\bm{x}_j$ are the $d$-dimensional feature vectors, $\theta\left(\bm{p}_i, \bm{p}_j\right)\in\mathbb{R}^{d}$ and $\psi\left(\bm{x}_i, \bm{x}_j\right)\in\mathbb{R}^{d}$ are two embeddings in the spatial and feature spaces, respectively, $\bm{\tilde{x}}_{i,l}$ is the  $l$-th  element of the interpolated feature $\bm{\tilde{x}}_{i}$, $l=1,2,\cdots, d$. The embeddings $\theta\left(\bm{p}_i, \bm{p}_j\right)$ and $\psi\left(\bm{x}_i, \bm{x}_j\right)$ can be defined as:
\begin{equation}
\begin{aligned}
& \theta\left(\bm{p}_i, \bm{p}_j\right) = \operatorname{ReLU}(\bm{W}_{\theta, j}^{\top}\left(\bm{p}_i-\bm{p}_j\right)),\;\; \psi\left(\bm{x}_i, \bm{x}_j\right) = \operatorname{ReLU}(\bm{W}_{\psi, j}^{\top}\left(\bm{x}_i-\bm{x}_j\right))
\end{aligned}
\end{equation}
where $\operatorname{ReLU}$ is the activation function, $\bm{W}_{\theta,j}\in\mathbb{R}^{3\times d}$ and $\bm{W}_{\psi,j}\in\mathbb{R}^{d\times d}$ are the weights to be learned.  Based on the differences between the points $\bm{p}_i$ and $\bm{p}_j$ ,  $\bm{p}_i-\bm{p}_j$ and $\bm{x}_i-\bm{x}_j$,  the embeddings $\theta\left(\bm{p}_i, \bm{p}_j\right)$ and $\psi\left(\bm{x}_i, \bm{x}_j\right)$ can encode local structure information of the point $\bm{p}_i$  in the spatial and feature spaces, respectively. It is noted that in Eq.~\ref{equ:bilateral} the channel-wise bilateral interpolation is adopted. As shown in Fig.~\ref{fig:interpolation}, the new interpolated feature
$\bm{\tilde{x}}_{i}$ can be obtained from the neighborhood of $\bm{x}_i$ with the bilateral weight. For each point, we perform the bilateral interpolation in the  $k$-neighborhood to generate new $k$ points. Therefore, we can obtain a high-resolution feature map, where the neighborhood of each point contains 2$k$ points.

%\subsection{Fusion Network}
%\textbf{Feature Fusion}
After the interpolation, we then apply the convolution on the enlarged feature maps. For each point, we divide the neighborhood of $2k$ points into two regions according to the distance. As shown in Fig.~\ref{fig:deconv}~(c), the closest $k$ points belong to the first region and the rest as the second region. Similar to PointNet~\cite{qi2017pointnet}, we first use the multi-layer perceptron to generate high-dimensional feature maps and then use the max-pooling operation to obtain the local features of the two interpolated points from two regions. As shown in Fig.~\ref{fig:deconv}~(d), we can double the number of points from the inputs through the deconvolution network to generate a high-resolution local feature map $\bm{X}_{local}$.% In the experiments, we observe that global features are also very helpful to generate high quality point clouds.
We also use the max-pooling operation to extract the global feature of point clouds. By replicating the global feature for $N$ times, where $N$ is the number of points, we can obtain the high-resolution global feature map $\bm{X}_{global}$. Then we concatenate the local feature map $\bm{X}_{local}$ and the global feature map $\bm{X}_{global}$ to obtain the output of the deconvolution network $\bm{X}_{c} = \left[\bm{X}_{local}; \bm{X}_{global}\right]$. Thus, the output $\bm{X}_{c}$ can not only characterize the local geometric structures of point clouds, but also capture the global shape of point clouds during the point cloud generation process.

%\subsection{Generation Block}
\textbf{3D point cloud generation.} Our goal is to progressively generate 3D point clouds from the low resolution to the high resolution. Stacked deconvolution networks can progressively double the number of points and generate their high-dimensional feature maps.
As shown in Fig.~\ref{fig:network_generator}, we use the MLP after each deconvolution network to generate the 3D coordinates of point clouds at each resolution. Note that two outputs of the DECONV block are the same, one for generating 3D coordinates of point clouds and the other as the features of the point clouds. We concatenate the generated 3D coordinates with the corresponding features as an input to the next DECONV block.

In our framework, we employ the PointNet~\cite{qi2017pointnet} as our discriminator. Generated point clouds from the progressive deconvolution generator are fed into the discriminator to distinguish whether the generated point clouds are from the real point clouds. For different resolutions, the network parameters of the discriminators are different.

\subsection{Shape-preserving adversarial loss}
In this subsection, in order to ensure that the geometric structures of the generated point clouds are consistent between different resolutions, we impose a shape constraint on the generators to formulate a shape-preserving adversarial loss.

\textbf{Shape-consistent constraint.} During the training process, different resolutions of 3D point clouds are generated. With the increase of the resolution of the output of the progressive deconvolution network, generated point clouds become more and more denser. It is expected that the local geometric structures of the generated point clouds are as consistent as possible between different resolutions. Since our progressive deconvolution generation network is an unsupervised generation model, it is difficult to distinguish different shapes from the same class of 3D objects for the discriminator. Thus, for the specific class of 3D objects, the deconvolution generation networks at different resolutions might generate 3D point clouds with different shapes. Therefore, we encourage that the means and covariances of the neighborhoods of the corresponding points between different resolutions are as close as possible so that the corresponding parts of different resolutions of generated point clouds are consistent.    %We can find that the coordinate mean, variance and covariance of the generated point cloud are independent of the number of point clouds. Therefore, we assume that the coordinates of different resolution point clouds with the same shape should have the same mean, variance at corresponding parts. Specifically, as shown in Fig.~\ref{fig:spl}, we take three query balls in the corresponding position, each of which represents a local area. For example, blue query balls are placed in the corresponding area on the legs of two tables with different resolutions. Although the resolution is different, the geometry of the table legs is consistent. The mean and variance of the local region can describe the local geometric characteristics to a certain extent. To keep the geometry consistent, the mean and variance of the table legs in the corresponding areas should be similar to each other. When the geometry of the corresponding parts of two objects with different resolutions remains the same, the two objects also remain the same shape.

\textbf{Shape-preserving adversarial loss.} We employ the mean and covariance of the neighborhoods of the corresponding points to characterize the consistency of the local geometric structures of the generated point clouds between different resolutions. We use the farthest point sampling (FPS) to choose centroid points from each resolution and find the $k$-neighborhood for centroid points. The mean and covariance of the neighborhood of the $i$-th centroid point are represented as:
\begin{equation}
\bm{\mu}_i = \frac{\sum\nolimits_{j\in\mathcal{N}_i}\bm{p}_j}{k} ,\;\; \bm{\sigma}_i = \frac{{\sum\nolimits_{j\in\mathcal{N}_i}\left(\bm{p}_j-\bm{\mu}_i\right)^{\top}\left(\bm{p}_j-\bm{\mu}_i\right)}}{k-1}
\end{equation}
where $\mathcal{N}_i$ is the neighborhood of the centroid point, $\bm{p}_j \in\mathbb{R}^{3}$ is the coordinates of the point cloud, $\bm{\mu}_i \in\mathbb{R}^{3}$ and $\bm{\sigma}_i \in\mathbb{R}^{3\times3}$ are the mean and covariance of the neighborhood, respectively. %Each $k$-neighborhood represent a local path. Therefore, we

%Similarly, we use the FPS to sample $M_i$ centroids points from the ($i$+1)-th resolution.
%We aim to minimize the mean and covariance of the $k$-neighborhood of the corresponding centroid points between adjacent resolutions.
Since the sampled centroid points are not completely matched between adjacent resolutions, we employ the Chamfer distances of the means and covariances to formulate the shape-preserving loss. We denote the centroid point sets at the resolutions $l$ and $l+1$ by $S_{l}$ and $S_{l+1}$, respectively. %We first concatenate the mean and covariance of the $i$-th centroid point as:
%\begin{equation}
%\bm{w}_{i}=[\bm{\mu}_{i},\; \operatorname{Resize}(\bm{\sigma}_{i})]
%\end{equation}
%where $[\cdot,\cdot]$ denotes the concatenation and $\bm{w}_{i} \in\mathbb{R}^{12}$. Note that $\bm{\sigma}_{i}\in\mathbb{R}^{3\times3}$ is resized as $\bm{\sigma}_{i}\in\mathbb{R}^{9}$.
The Chamfer distance $d_{1}(S_{l}, S_{l+1})$ between the means of the neighborhoods from the adjacent resolutions is defined as:
%\begin{equation}
%\begin{aligned}
%d_{CH}(S,\widehat{S})=\max \left\lbrace \frac{1}{|S|}\sum_{x\in %S}\min_{\widehat{x} \in \widehat{S}}\|x-\widehat{x}\|_2,
%\frac{1}{|\widehat{S}|}\sum_{\widehat{x}\in \widehat{S}}\min_{x \in %S}\|\widehat{x}-x\|_2 \right\rbrace
%\end{aligned}
%\end{equation}
\begin{equation}
\small
\begin{aligned}
d_{1}(S_{l}, S_{l+1})=\max &\left\{\frac{1}{|S_{l}|} \sum_{i \in S_{l}} \min _{j \in S_{l+1}}\|\bm{\mu}_{i}-\bm{\mu}_{j}\|_{2}\right.,%\\
&\left.\frac{1}{|S_{l+1}|} \sum_{j \in S_{l+1}} \min _{i \in S_{l}}\|\bm{\mu}_{j}-\bm{\mu}_{i}\|_{2}\right\}
\end{aligned}
\end{equation}
Similarly, the Chamfer distance $d_{2}(S_{l}, S_{l+1})$ between the covariances of the neighborhoods is defined as:
\begin{equation}
\small
\begin{aligned}
d_{2}(S_{l}, S_{l+1})=\max &\left\{\frac{1}{|S_{l}|} \sum_{i \in S_{l}} \min _{j \in S_{l+1}}\|\bm{\sigma}_{i}-\bm{\sigma}_{j}\|_{F}\right.,%\\
&\left.\frac{1}{|S_{l+1}|} \sum_{j \in S_{l+1}} \min _{i \in S_{l}}\|\bm{\sigma}_{j}-\bm{\sigma}_{i}\|_{F}\right\}
\end{aligned}
\end{equation}
%where
The shape-preserving loss (SPL) for multi-resolution point clouds is formulated as:
\begin{equation}
SPL(G_l, G_{l+1}) =\sum\nolimits_{l=1}^{M-1}d_{1}(S_{l}, S_{l+1})+d_{2}(S_{l}, S_{l+1})
\label{equ:spl_loss}
\end{equation}
where $M$ is the number of resolutions, $G_l$ and $G_{l+1}$ represents the $l$-th and $(l+1)$-th point cloud generators, respectively. %In this way, we can keep the local geometry consistent in the corresponding local area so that the overall shape remains consistent.

Based on Eq.~\ref{equ:spl_loss}, for the generator $G_l$ and discriminator $D_l$, we define the following shape-preserving adversarial loss:
\begin{equation}
\begin{small}
\begin{aligned}
& L(D_l)=E_{\bm{s} \sim p_{real}(\bm{s})}(\log D_l (\bm{s}) + \log (1 - D_l (G_l (\bm{z}))))\\
& L(G_l)=E_{\bm{z} \sim p_{\bm{z}}(\bm{z})} (\log (1 - D_l(G_l (\bm{z}))))+\lambda SPL (G_l(\bm{z}), G_{l+1}(\bm{z}))
\end{aligned}
\end{small}
\end{equation}
where $\bm{s}$ is the real point cloud sample, $\bm{z}$ is the randomly sampled latent vector from the distribution $p(\bm{z})$ and $\lambda$ is the regularization parameter. Note that we ignore the SPL in $L(G_l)$ for $l=M$.
Thus, multiple generators $G$ and discriminators $D$ can be trained with the following equation:
\begin{equation}
max_{D}\sum\nolimits_{l=1}^{M}L(D_l),
min_{G}\sum\nolimits_{l=1}^{M}L(G_l)
\end{equation}
where $D=\{D_1, D_2, \cdots, D_{M}\}$ and $G=\{G_1, G_2, \cdots, G_{M}\}$. During the training process, multiple generators $G$ and discriminators $D$ are alternatively optimized till convergence.

%----------------------------------------------------

\section{Experiments}
In this section, we first introduce the experimental settings. We then compare our method to state-of-the-art point cloud generation methods. %in terms of the quantitative evaluation. %We then compare our method to state-of-the-art point cloud generation methods in terms of visual results and quantitative evaluation. 
Finally, we analyze the effectiveness of our proposed point cloud generation method.

\subsection{Experimental Settings}
We evaluate our proposed generation network on three popular datasets including ShapeNet~\cite{chang2015shapenet}, ModelNet10 and ModelNet40 ~\cite{wu20153d}. ShapeNet is a richly annotated large-scale point cloud dataset containing 55 common object categories and 513,000 unique 3D models. In our experiments, we only use 16 categories of 3D objects.  ModelNet10 and ModelNet40 are subsets of ModelNet, which contain 10 categories and 40 categories of CAD models, respectively.

Our proposed framework mainly consists of progressive deconvolution generator and shape-preserving discriminator. In this paper, we generate four resolutions point cloud from a 128-dimensional latent vector. In the generator, the output size of 4 deconvolution networks are 256$\times$32, 512$\times$64, 1024$\times$128 and 2048$\times$256. To generate point clouds, we use 4 MLPs with the same settings. Note that MLPs are not shared for 4 resolutions. After the MLP, we adopt the $Tanh$ activation function. In the discriminator, we use 4 PointNet-like structures. For different resolutions, the network parameters of the discriminators are different. We use Leaky ReLU~\cite{xu2015empirical} and batch normalization~\cite{ioffe2015batch} after every layer. The more detailed structure of our framework is shown in the Appendix. In addition, we use the $k=20$ nearest points as the neighborhood for the bilateral interpolation. During the training process, we adopt Adam~\cite{kingma2014adam} with the learning rate $10^{-4}$ for both generator and discriminator. We employ an alternative training strategy in ~\cite{goodfellow2014generative} to train the generator and discriminator. Specifically, the discriminator is optimized for each generator step. %In addition, the proposed GAN-based model is easy to train and stable. More discussion is listed in the Appendix.
Furthermore, we observe that our GAN-based model is easy to train and stable during training. More analysis is shown in the Appendix.

\subsection{Evaluation of point cloud generation}
%To validate the effectiveness of our proposed point cloud generation method, we perform qualitative and quantitative comparisons with the state-of-the-art methods. We first visualize the generated point clouds and then present the quantitative performance comparison.
In this subsection, we first visualize the generated point clouds. Then we present the quantitative performance comparison.

\begin{figure}
	\centering
	\includegraphics[width=0.88\linewidth]{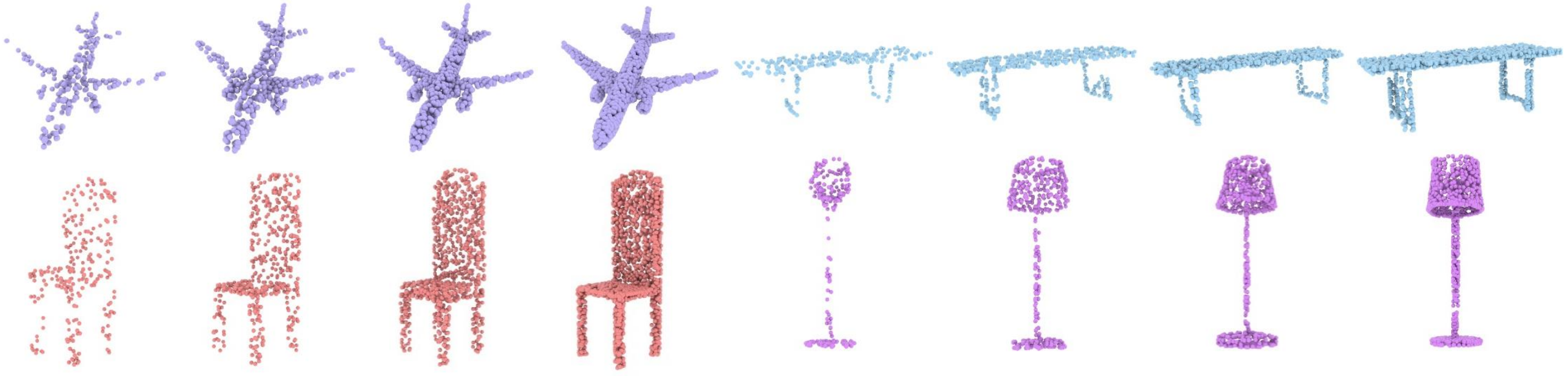}
	\caption{Generated point clouds including ``Airplane'', ``Table'', ``Chair'' and ``Lamp''. Each category has four resolutions of point clouds (256, 512, 1024 and 2048).%Note that the four resolutions point clouds of the same category are generated from the same latent vector.
	}
	\label{fig:visualization}
\end{figure}

\textbf{Visual results.} As shown in Fig.~\ref{fig:visualization}, on the ShapeNet~\cite{chang2015shapenet} dataset, we visualize the synthesized point clouds containing 4 categories, which are ``Airplane'', ``Table'', ``Chair'',  and ``Lamp'', respectively. Due to our progressive generator, each category contains four resolutions of point clouds (256, 512, 1024 and 2048) generated from the same latent vector. It can be observed that the geometric structures of different resolutions of generated point clouds are consistent. Note that the generated point clouds contain detailed structures, which are consistent with those of real 3D objects. More visualizations are shown in the supplementary material.

\textbf{Quantitative evaluation.} To conduct a quantitative evaluation of the generated point clouds, we adopt the evaluation metric proposed in~\cite{achlioptas2017learning,LopezPaz2016RevisitingCT}, including Jensen-Shannon Divergence (JSD), Minimum Matching Distance (MMD) and Coverage (COV), the earth mover's distance (EMD), the chamfer distance (CD) and the 1-nearest neighbor accuracy (1-NNA). JSD measures the marginal distributions between the generated samples and  real samples. MMD is the distance between one point cloud in the real sample set and its nearest neighbors in the generation set. COV measures the fraction of point clouds in the real sample set that can be matched at least one point cloud in the generation set. 1-NNA is used as a metric to evaluate whether two distributions are identical for two-sample tests. Table.~\ref{tab:metric} lists our results with different criteria on the ``Airplane'' and ``Chair'' categories in the ShapeNet dataset. In Table.~\ref{tab:metric},  except for PointFlow~\cite{pointflow} (VAE-based generation method), the others are GAN-based generation methods. For these evaluation metrics, in most cases, our point cloud deconvolution generation network (PDGN)  outperforms other methods, demonstrating the effectiveness of the proposed method. Moreover, the metric results on the ``Car'' category and the mean result of all 16 categories are shown in the supplementary material.

\begin{table}
	\caption{The results on the ``Airplane'' and ``Chair'' categories.  Note that JSD scores and MMD-EMD scores are multiplied by 10$^2$. MMD-CD scores are multiplied by 10$^3$. Lower JSD, MMD-CD, MMD-EMD, 1-NNA-CD, and 1-NNA-EMD show better performance while higher COV-CD and COV-EMD indicate better performance.
	}
	\begin{center}
		\resizebox{0.8\textwidth}{!}
		{ 
			\begin{tabular}{llcccccccccc}
				\toprule
				\multicolumn{1}{c}{\multirow{2}{*}{Category}} &\multicolumn{1}{c}{\multirow{2}{*}{Model}}&  \multicolumn{1}{c}{\multirow{2}{*}{JSD ($\downarrow$)\;\;\;\;}} & \multicolumn{2}{c}{MMD ($\downarrow$)} & & \multicolumn{2}{c}{COV (\%, $\uparrow$)} & & \multicolumn{2}{c}{1-NNA (\%, $\downarrow$)} \\
				\cmidrule{4-5} \cmidrule{7-8} \cmidrule{10-11}
				&\multicolumn{1}{c}{}&&CD&EMD&&CD&EMD&&CD&EMD\\
				\midrule
				\multirow{8}{*}{Airplane}&r-GAN~\small\cite{achlioptas2017learning}&7.44&0.261&5.47&&42.72&18.02&&93.50&99.51 \\
				&l-GAN (CD)~\small\cite{achlioptas2017learning}&4.62&0.239&4.27&&43.21&21.23&&86.30&97.28\\
				&l-GAN (EMD)~\small\cite{achlioptas2017learning}&3.61&0.269&3.29&&47.90& 50.62&&87.65&85.68\\
				&PC-GAN~\small\cite{Li2018PointCG}&4.63&0.287&3.57&&36.46&40.94&&94.35&92.32\\
				&GCN-GAN~\small\cite{valsesia2018learning} &8.30&0.800&7.10&&31.00&14.00&&-&-\\
				&tree-GAN~\small\cite{shu20193dpc}&9.70&0.400&6.80&&61.00&20.00&&-&-\\
				&PointFlow~\small\cite{pointflow}&4.92&{\bf 0.217}&3.24&&46.91&48.40&&75.68& 75.06\\
				&PDGN (ours)&{\bf 3.32}&0.281&{\bf 2.91}&&{\bf 64.98} &{\bf 53.34}&&{\bf 63.15}&{\bf 60.52}&\\
				\midrule
				
				\multirow{8}{*}{Chair}&r-GAN~\small\cite{achlioptas2017learning}&11.5&2.57&12.8&&33.99&9.97&&71.75&99.47 \\
				&l-GAN (CD)~\small\cite{achlioptas2017learning}&4.59&2.46&8.91&&41.39&25.68&&64.43&85.27\\
				&l-GAN (EMD)~\small\cite{achlioptas2017learning}&2.27&2.61&7.85&&40.79&41.69&&64.73&65.56\\
				&PC-GAN~\small\cite{Li2018PointCG}&3.90&2.75&8.20&&36.50&38.98&&76.03&78.37\\
				&GCN-GAN~\small\cite{valsesia2018learning}&10.0&2.90&9.70&&30.00&26.00&&-&-\\
				&tree-GAN~\small\cite{shu20193dpc}&11.9&{\bf 1.60}&10.1&&58.00&30.00&&-&-\\
				&PointFlow~\small\cite{pointflow}&1.74&2.24&7.87&&46.83&46.98&& 60.88& 59.89\\
				&PDGN (ours) &{\bf 1.71}&1.93&{\bf 6.37}&&{\bf 61.90}&{\bf 57.89}&&{\bf 52.38}& {\bf 57.14}\\
				
				\bottomrule
			\end{tabular}
		}
	\end{center}
	\label{tab:metric}
\end{table}

\begin{figure}
	\begin{center}
		\includegraphics[width=0.85\linewidth]{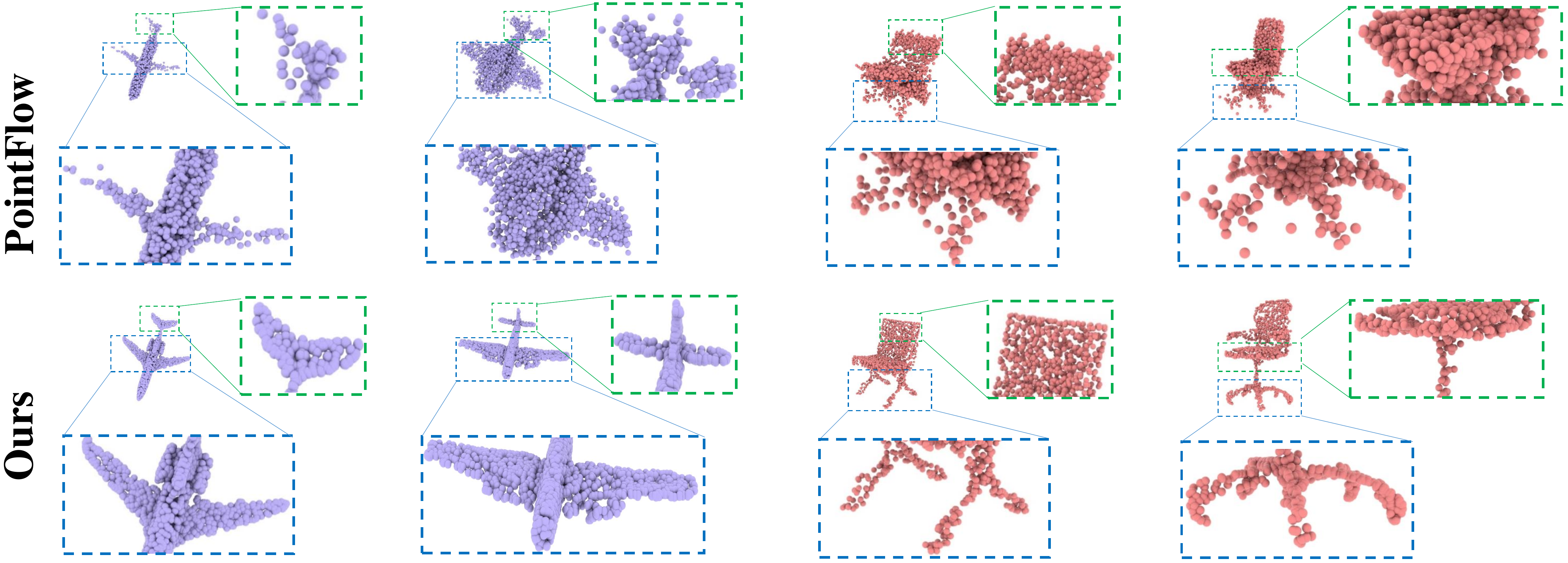}
	\end{center}
	\caption{Visualization results of our method and PointFlow on the ``Airplane'' and ``Chair'' categories.%, respectively. Our method can generate more complete point clouds. Especially, our method successfully addresses the cases with many thin structures such as the ``Chair'' category%The three views from left to right are the main view, left view, and top view.
	}
	\label{fig:cmp_pointflow}
\end{figure}

Different from the existing GAN-based generation methods, we develop a progressive generation network to generate multi-resolution point clouds. In order to generate the high-resolution point clouds, we employ the bilateral interpolation in the spatial and feature spaces of the low-resolution point clouds to produce the geometric structures of the high-resolution point clouds. Thus, with the increase of resolutions, the structures of generated point clouds are more and more clear. Therefore, our PDGN can yield better performance in terms of these evaluation criteria. In addition, compared to PointFlow, our method can perform better on point clouds with thin structures. As shown in Fig.~\ref{fig:cmp_pointflow}, it can be seen that our method can generate more complete point clouds. %Especially, our method successfully addresses the cases with many thin structures such as the ``Chair'' category. 
Since in PointFlow the VAE aims to minimize the lower bound of the log-likelihood of the latent vector, it may fail for point clouds with thin structures. Nonetheless, due to the bilateral deconvolution and progressive generation from the low resolution to the high resolution, our PDGN can still achieve good performance for point cloud generation with thin structures. For more visualization comparisons to PointFlow please refer to the supplementary material.

\textbf{Classification results.} Following ~\cite{wu2016learning,pointflow}, we also conduct the classification experiments on ModelNet10 and ModelNet40 to evaluate our generated point clouds. We first use all samples from ModelNet40 to train our network with the iteration of 300 epochs. Then we feed all samples from ModelNet40 to the trained discriminator (PointNet) for feature extraction. With these features, we simply train a linear SVM to classify the generated point clouds. The settings of ModelNet10 are consistent with ModelNet40.  The classification results are listed in Table.~\ref{tab:classification}.  Note that for a fair comparison we only compare the point cloud generation methods in the classification experiment.
%The detailed results are listed in the Appendix. %Note that the reported results include only the results of the point cloud generation method, not the results specific to the point cloud processing method.
%The classification results are 94.2\% on ModelNet10 dataset and 87.3\% on ModelNet40 dataset.
It can be found that our PDGN outperforms the state-of-the-art point cloud generation methods on the ModelNet10 and ModelNet40 datasets. The results indicate that the discriminator in our framework can extract discriminative features. Thus, our generator can produce high-quality 3D point clouds.

\begin{table}
	\caption{Classification results of various methods on ModelNet10 (MN10) and ModelNet40 (MN40) datasets. %Please note that the results of all reports are based on unsupervised methods. The better results are highlighted in \textbf{bold}.
	}
	\begin{center}
		\resizebox{0.44\textwidth}{!}
		{
			\begin{tabular}{lcc}
				\toprule
				\multicolumn{1}{c}{Model} & MN10 (\%) & MN40 (\%) \\
				\midrule
				SPH~\cite{kazhdan2003rotation} & 79.8 & 68.2 \\
				LFD~\cite{chen2003visual} & 79.9 & 75.5\\
				T-L Network~\cite{girdhar2016learning} & - & 74.4\\
				VConv-DAE~\cite{sharma2016vconv} & 80.5 & 75.5\\
				3D-GAN~\cite{wu2016learning} & 91.0 & 83.3\\
				PointGrow~\cite{Sun2018PointGrowAL} & - & 85.7\\
				MRTNet~\cite{gadelha2018multiresolution} & 91.7 & 86.4 \\
				PointFlow~\cite{pointflow} & 93.7 & 86.8 \\
				PDGN (ours) & {\bf 94.2} & {\bf 87.3}\\
				\bottomrule
			\end{tabular}
		}
	\end{center}
	\label{tab:classification}
\end{table}

\textbf{Computational cost.} We compare our proposed method to PointFlow and tree-GAN in terms of the training time and GPU memory. We conduct point cloud generation experiments on the ``Airplane'' category in the ShapeNet dataset. For a fair comparison, both codes are run on a single Tesla P40 GPU using the PyTorch~\cite{paszke2019pytorch} framework. For training 1000 iterators with 2416 samples of the ``Airplane'' category, our proposed method costs about 1.9 days and 15G GPU memory, while PointFlow costs about 4.5 days and 7.9G GPU memory, and tree-GAN costs about 2.5 days and 9.2G GPU memory. Our GPU memory is larger than others due to the four discriminators.

\subsection{Ablation study and analysis}
%In this subsection, we conduct ablation studies to demonstrate the effectiveness of the proposed method. We also compare our method with other methods.

\begin{figure}
	\begin{center}
		\includegraphics[width=0.95\linewidth]{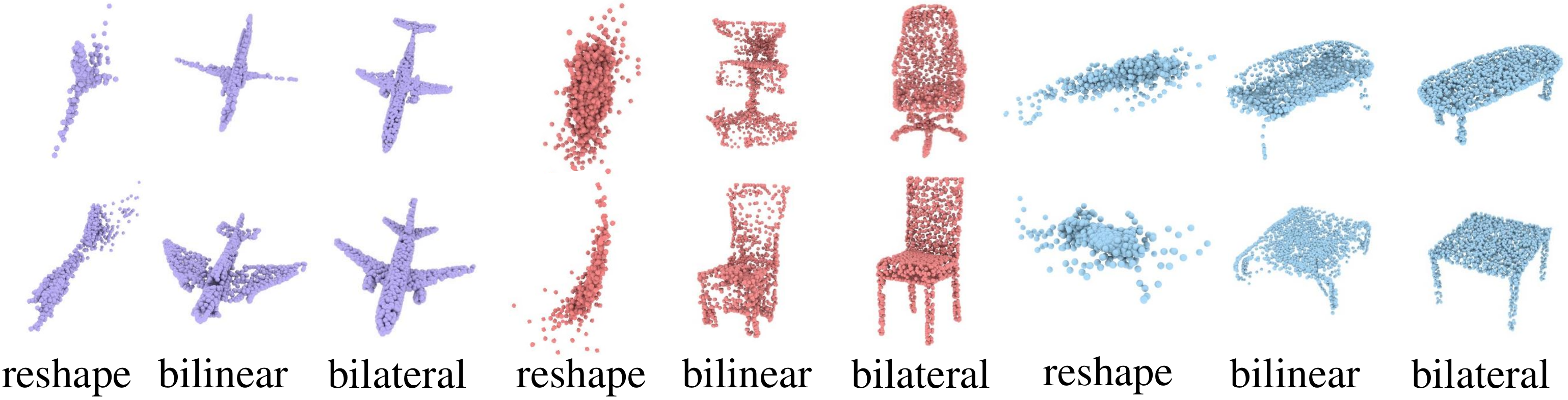}
	\end{center}
	\caption{Visualization results with different operations in the deconvolution network for three categories of point clouds.}
	\label{fig:ablation_bilateral_interpolation}
\end{figure}

\textbf{Bilateral interpolation.} In this ablation study, we conduct the experiments with different ways to generate the high-resolution feature maps, including the conventional reshape operation, bilinear interpolation and learning-based bilateral interpolation. In the conventional reshape operation,  we resize the feature maps to generate new points. As shown in Fig.~\ref{fig:ablation_bilateral_interpolation}, we visualize the generated point clouds from different categories. From the visualization results, one can see that the learning-based bilateral interpolation can generate more realistic objects than the other  methods. For example, for the ``Table'' category, with the learning-based bilateral interpolation, the table legs are clearly generated. On the contrary, with the bilinear interpolation and reshape operation, the generated table legs are not complete. Besides, we also conduct a quantitative evaluation of generated point clouds. As shown in Table.~\ref{tab:ablation_metric}, on the ``Chair'' category,   PDGN with the bilateral interpolation can obtain the best metric results. In contrast to the bilinear interpolation and reshape operation, the learning-based bilateral interpolation exploits the spatial coordinates and high-dimensional features of the neighboring points to adaptively learn weights for different classes of 3D objects. Thus, the  learned weights in the spatial and feature spaces can characterize the geometric structures of point clouds better. Therefore, the bilateral interpolation can yield good performance.

\begin{table}
	\caption{The ablation study results on the ``Chair'' category. %The best results are highlighted in \textbf{bold}. PDGN (bilateral interpolation), PDGN (Shape-preserving loss), PDGN ($k$=20) and PDGN (2048 points) are the same models
	}
	\begin{center}
		\resizebox{0.9\textwidth}{!}
		{ 
			\begin{tabular}{lcccccccccc}
				\toprule
				\multicolumn{1}{c}{\multirow{2}{*}{Model}}&  \multicolumn{1}{c}{\multirow{2}{*}{JSD ($\downarrow$)\;\;\;\;}} & \multicolumn{2}{c}{MMD ($\downarrow$)} & & \multicolumn{2}{c}{COV (\%, $\uparrow$)} & & \multicolumn{2}{c}{1-NNA (\%, $\downarrow$)} \\
				\cmidrule{3-4} \cmidrule{6-7} \cmidrule{9-10}
				%\midrule
				%-&-&+&CD&EMD&+&CD&EMD&+&CD&EMD\\
				&\multicolumn{1}{c}{}&CD&EMD&\;\;&CD&EMD&\;\;&CD&EMD\\
				\midrule
				
				PDGN (reshape) &8.69&3.38&9.30&&55.01&44.49&&82.60& 80.43\\
				PDGN (bilinear interpolation) &5.02&3.31&8.83&&53.84&48.35&&69.23&68.18\\
				PDGN (bilateral interpolation) &{\bf 1.71}&{\bf 1.93}&{\bf 6.37}&&{\bf 61.90}&{\bf 57.89}&&{\bf 52.38}& {\bf 57.14}\\
				\midrule
				
				PDGN (adversarial loss) &3.28&3.00&8.82&&56.15&53.84&&57.14&66.07\\
				PDGN (EMD loss) &3.35&3.03 &8.80&&53.84&53.34&&60.89& 68.18\\
				PDGN (CD loss) &3.34&3.38&9.53&&55.88&52.63&&59.52& 67.65\\
				PDGN (shape-preserving loss) &{\bf 1.71}&{\bf 1.93}&{\bf 6.37}&&{\bf 61.90}&{\bf 57.89}&&{\bf 52.38}& {\bf 57.14}\\
				%\midrule
				
				%PDGN ($k$=4) &11.4&36.9&25.7&&22.58&28.57&&90.35&94.97\\
				%PDGN ($k$=12) &3.95&2.89&10.67&&45.83&47.79&&67.50&64.70\\
				%PDGN ($k$=20) &{\bf 1.71}&{\bf 1.93}&{\bf 6.37}&&61.90&{\bf 57.89}&&{\bf 52.38}& {\bf 57.14}\\
				%PDGN ($k$=28) &2.19&2.47&9.37&&55.88&52.94&&63.64&58.52\\
				%PDGN ($k$=36) &1.76&2.53&9.29&&{\bf 62.07}&57.13&&54.56&57.89\\
				
				\midrule
				
				PDGN (256 points) &5.57&5.12&9.69&&39.47&42.85&&67.56&70.27\\
				PDGN (512 points) &4.67&4.89&9.67&&47.82&51.17&&71.42&67.86\\
				PDGN (1024 points) &2.18&4.53&11.0&&56.45&55.46&&64.71&70.58\\
				PDGN (2048 points) &{\bf 1.71}&{\bf 1.93}&{\bf 6.37}&&{\bf 61.90}&{\bf 57.89}&&{\bf 52.38}& {\bf 57.14}\\
				
				\bottomrule
			\end{tabular}
		}
	\end{center}
	\label{tab:ablation_metric}
\end{table}

\begin{figure}
	\begin{center}
		\includegraphics[width=0.85\linewidth]{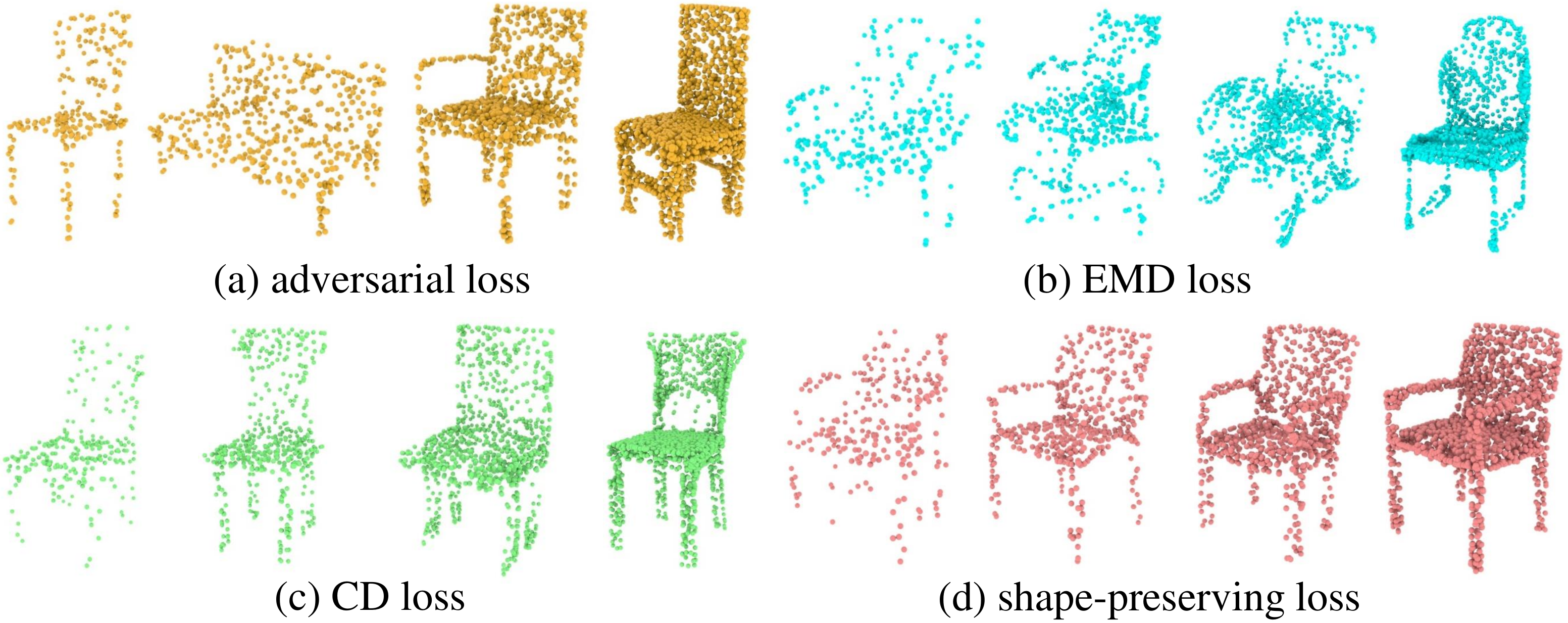}
	\end{center}
	\caption{Visualization results of generated point clouds with different loss functions. For each loss, four resolutions of point clouds (256, 512, 1024 and 2048) are visualized.}
	\label{fig:ablation_spl}
\end{figure}

\textbf{Shape-preserving adversarial loss.} To demonstrate the effectiveness of our shape-preserving adversarial loss, we train our generation model with the classical adversarial loss, EMD loss, CD loss and shape-preserving loss. It is noted that in the EMD loss and CD loss we replace the shape-preserving constraint (Eq.~\ref{equ:spl_loss}) with the Earth mover's distance and  Chamfer distance of point clouds between the adjacent resolutions, respectively.  We visualize the generated points with different loss functions in Fig.~\ref{fig:ablation_spl}. It can be found that  the geometric structures of different resolutions of generated point clouds are consistent with the shape-preserving adversarial loss. Without the shape-preserving constraint on the multiple generators, the classical adversarial loss cannot guarantee the consistency of generated points between different resolutions. Although the EMD or CD loss imposes the constraint on different resolutions of point clouds, the loss can only make the global structures of point clouds consistent. On the contrary, the shape-preserving loss can keep the consistency of the local geometric structures of multi-resolution point clouds with the mean and covariance of the neighborhoods. Thus, our method with the shape-preserving loss can generate high-quality point clouds. Furthermore, we also conduct a quantitative evaluation of generated point clouds. As shown in Table.~\ref{tab:ablation_metric}, metric results on the ``Chair'' category show that our method with the shape-preserving loss can obtain better results than the method with the other losses. %Note that results of EMD and CD losses are worse than others due to they cannot produce clear 3D shapes. 

\begin{figure}
	\begin{center}
		\includegraphics[width=0.4\linewidth]{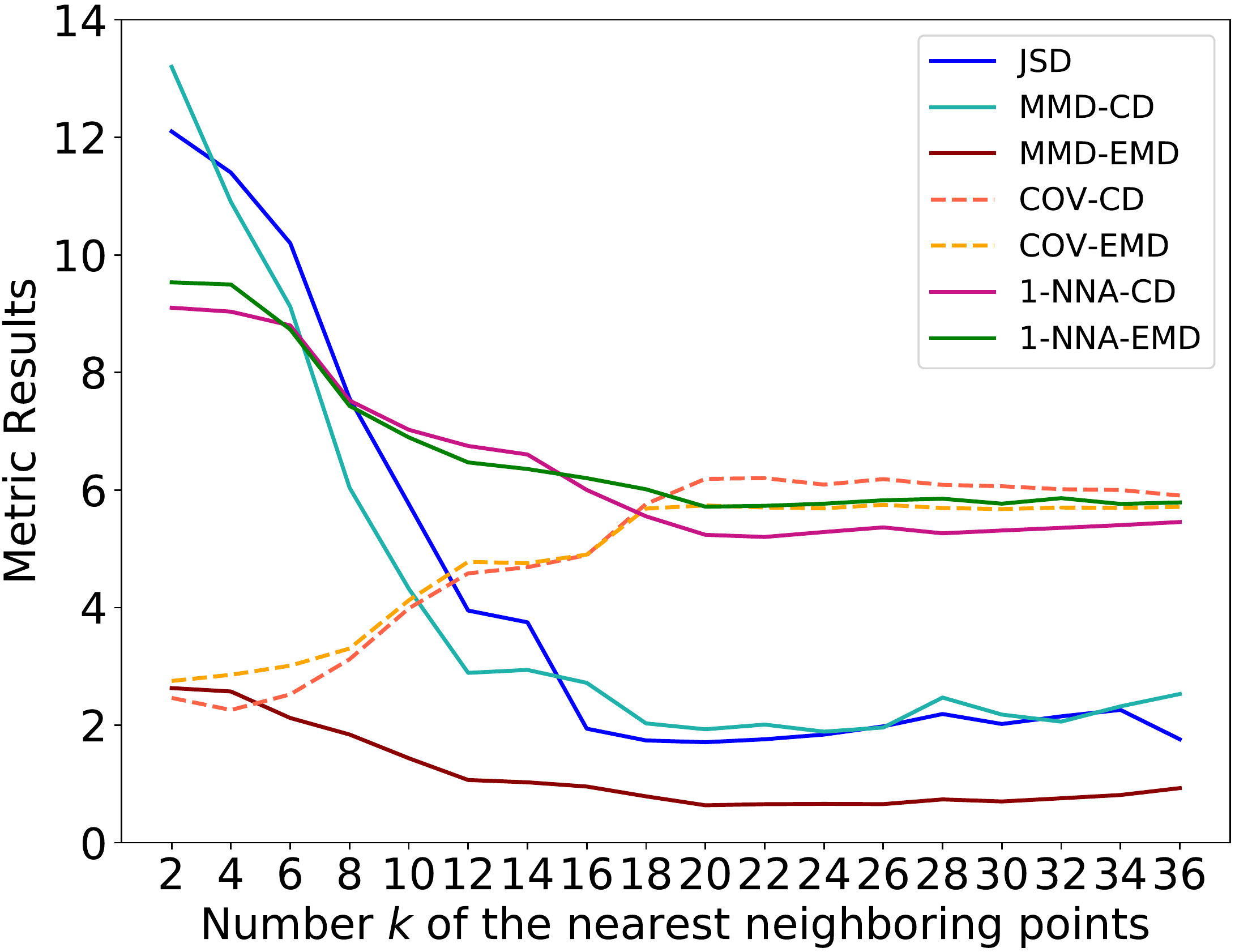}
	\end{center}
	\caption{Point cloud generation results with different metrics in the cases of different $k$ on the ``Chair'' category. Note that we magnify the results of COV-$*$ and 1-NNA-$*$ by a factor of 10.
	}
	\label{fig:ablation_k}
\end{figure}

\textbf{Parameter $k$.} To study the effect of parameter $k$ in the bilateral interpolation on the final generation result, we perform the ablation studies on parameter $k$. Specifically, $k$ represents the number of the nearest neighboring points in the bilateral interpolation. We select $k\in[2,4,\cdots,36]$ with interval 2. The metric results are shown in Fig.~\ref{fig:ablation_k}. It can be seen that setting $k$ values around 20 can obtain better performance than other choices. Actually, if $k$ is too small, the small neighborhood cannot produce the discriminative geometric features of the points, leading to the poor generation results. If $k$ is too large, the large neighborhood results in the high computational cost of the deconvolution operation. Therefore, for a good trade-off between the quality of generated point clouds and computational cost, we set $k$ to 20 in the experiments.

\textbf{Point cloud generation with different resolutions.} To verify the effectiveness of our progressive generation framework, we evaluate the metric results of generated point clouds in the cases of different resolutions. As shown in Table.~\ref{tab:ablation_metric}, on the ``Chair'' category, we report the results in the cases of four resolutions (256, 512, 1024 and 2048). One can see that as the resolution increases, the quality of the generated point clouds is gradually improved in terms of the evaluation criteria. In addition, as shown in Fig.~\ref{fig:visualization}, with the increase of resolutions, the local  structures of point clouds are also more and more clear.  This is because  our progressive generation framework can exploit the bilateral interpolation based deconvolution to generate the coarse-to-fine geometric structures of point clouds.

\section{Conclusions}
In this paper, we proposed a novel end-to-end generation model for point clouds. Specifically, we developed a progressive deconvolution network to generate multi-resolution point clouds from the latent vector. In the deconvolution network, we employed the learning-based bilateral interpolation to generate high-resolution feature maps so that the local structures of point clouds can be captured during the generation process. In order to keep the geometric structure of the generated point clouds at different resolutions consistent, we formulated the shape-preserving adversarial loss to train the point cloud deconvolution network. Experimental results on ShapeNet and ModelNet verify the effectiveness of our proposed progressive point cloud deconvolution network.

%------------------------------------------------------
\bibliographystyle{splncs04}
\bibliography{egbib}

\appendix
%Supplementary Material

\section{Overview}
This document provides additional technical details and more visualization results. Specifically, we first describe the details of the network architecture for the experiments in Sec.~\ref{sec:structure}. Then we discuss the training of the proposed GAN-based point cloud generation model in Sec.~\ref{sec:training}. Furthermore, we show more visual and quantitative results of our method in Sec.~\ref{sec:visualization}. Finally, in Sec.~\ref{sec:comparison}, we present more visualization results compared with PointFlow~\cite{pointflow}.

\section{Network Architecture}\label{sec:structure}
\begin{figure}
	\centering
	\includegraphics[width=0.65\linewidth]{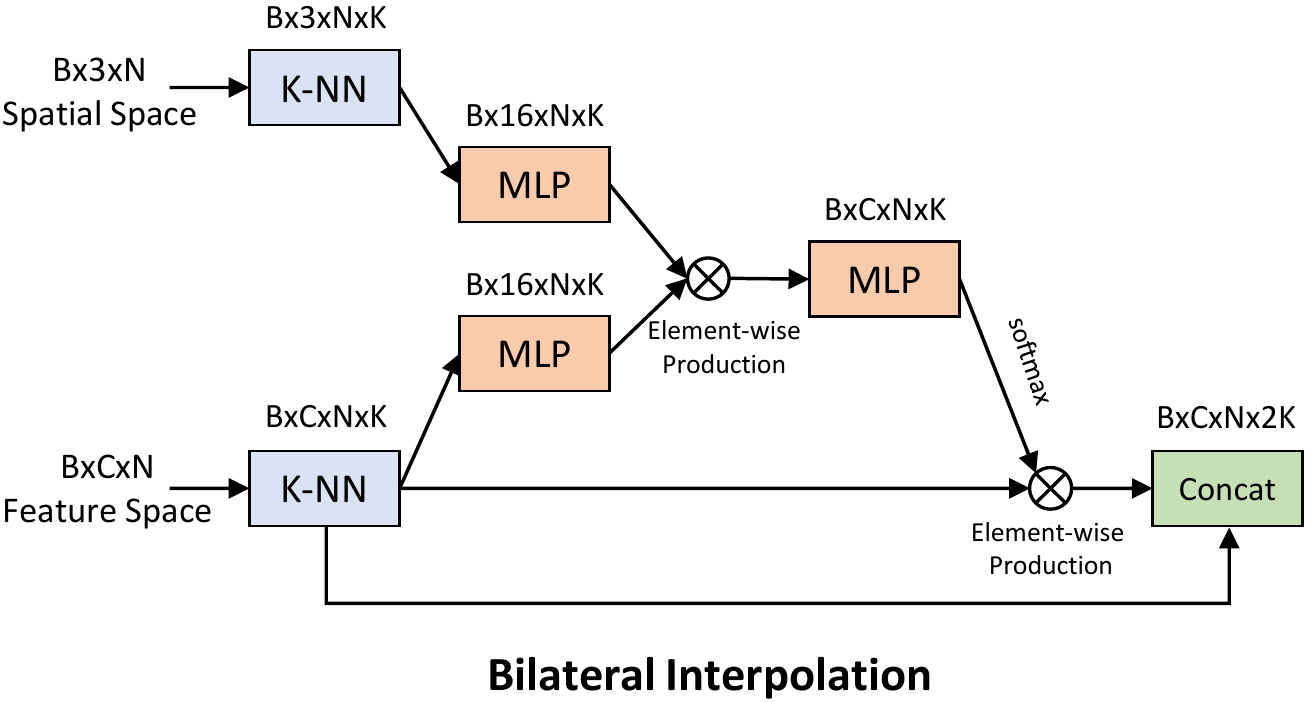}
	\caption{Learning-based bilateral interpolation in our experiments.}
	\label{fig:bilateral_interpolation}
\end{figure}

\begin{figure}
	\centering
	\includegraphics[width=0.9\linewidth]{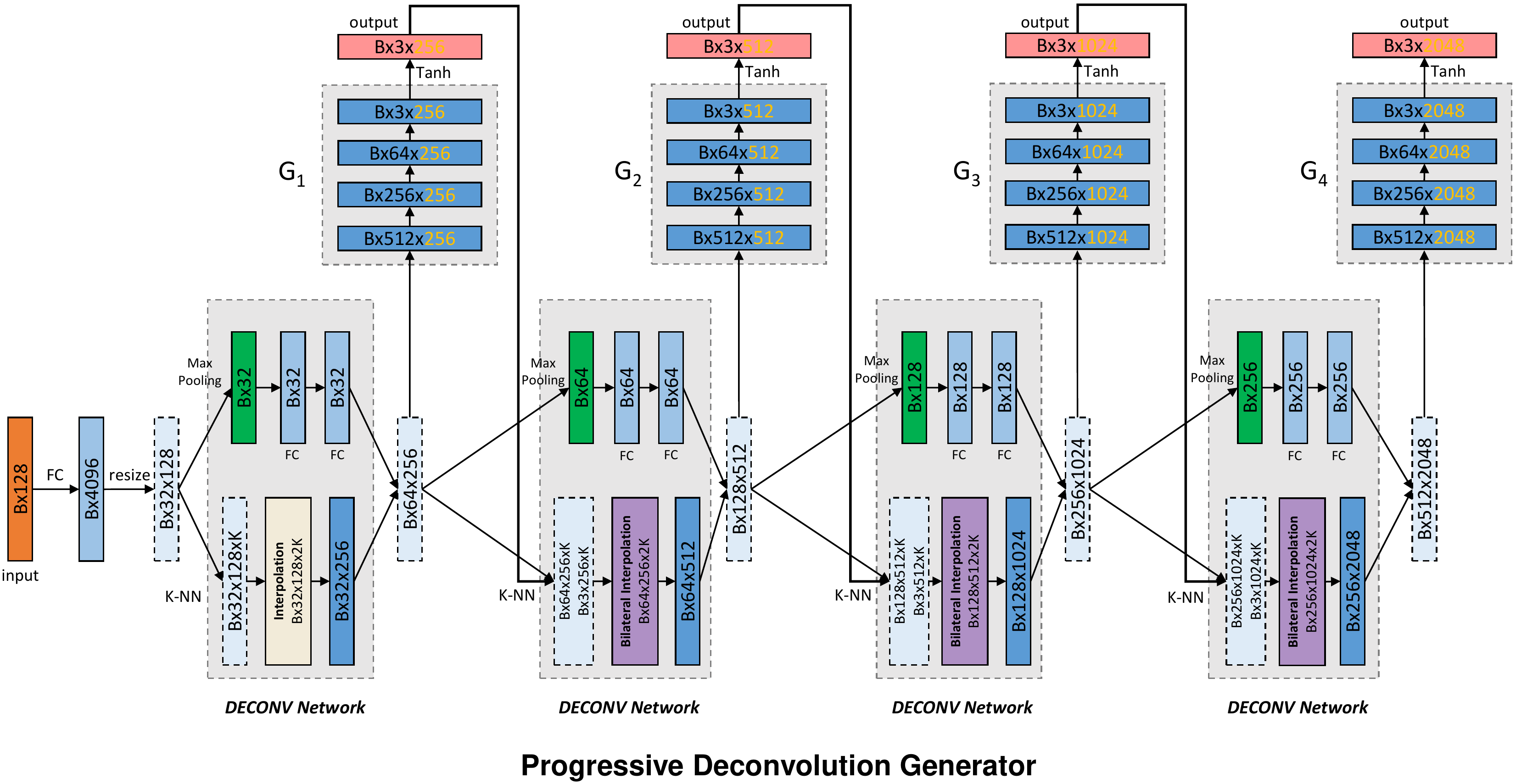}
	\caption{The architecture of our progressive deconvolution generator. $G_1$, $G_2$, $G_3$, and $G_4$ are four MLPs for 256, 512, 1024, and 2048 resolutions, respectively. FC is the fully connected layer. $k$-NN represents the $k$ nearest neighbor. }
	\label{fig:architecture_generator}
\end{figure}

For the generator, the structure is illustrated in Fig.~\ref{fig:architecture_generator}. We generate four resolutions (256, 512, 1024, and 2048) point cloud from a 128-dimensional latent vector creating by the normal distribution $\mathcal{N}(0,0.2)$. Specifically, we stack 4 deconvolution networks (DECONV Network) to generate multi-resolution feature maps. Each deconvolution network has two branches: one for capturing global information and one for bilateral interpolation. For bilateral interpolation, the details structure is shown in Fig.~\ref{fig:bilateral_interpolation}.The output size of 4 deconvolution networks are 256$\times$32, 512$\times$64, 1024$\times$128, and 2018$\times$256, respectively. It is important to note that our four deconvolution networks are not shared at four resolutions. To generate 3D point cloud coordinates, after each deconvolution network, we use Multi-Layer Perceptrons (MLPs) with neuron sizes of 512, 256, 64, and 3, respectively. After MLPs, we use $Tanh$ as the activation function to generate the final 3D coordinates.

\begin{figure}
	\centering
	\includegraphics[width=0.9\linewidth]{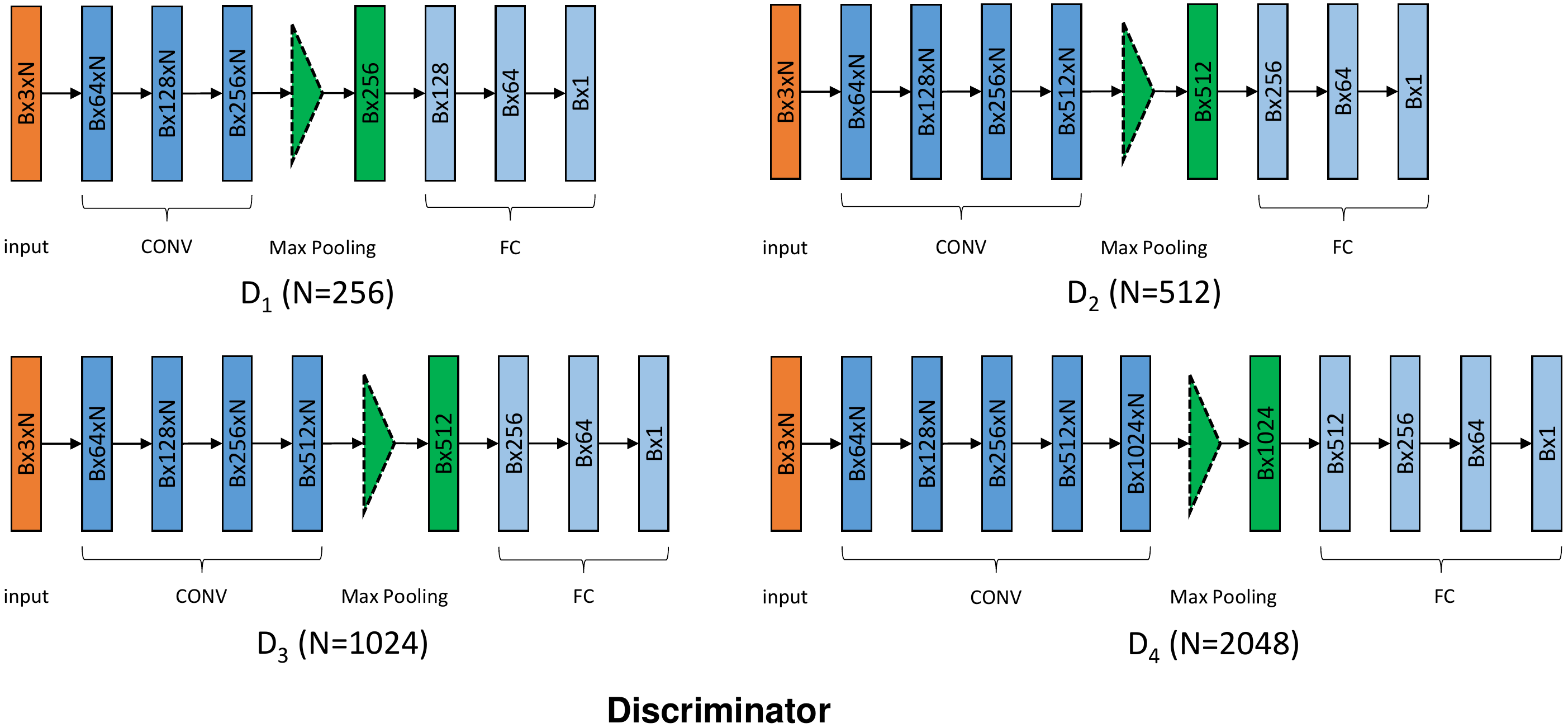}
	\caption{The architecture of our discriminators. $D_1$, $D_2$, $D_3$, and $D_4$ for resolution 256, 512, 1024, and 2048, respectively. FC is the fully connected layer. CONV represents 1$\times$1 convolution operating on each point.}
	\label{fig:architecture_discriminator}
\end{figure}

For the discriminator, in Fig.~\ref{fig:architecture_discriminator}, we adopt four PointNet-like~\cite{qi2017pointnet} networks as our discriminators. Specifically, we modify the network to accommodate different resolutions. In the experiments, we found that too many convolution layers are harmful for low-resolution point clouds. Besides, for different resolutions, the discriminators are not shared.

\section{Training of the Proposed GAN-based Model}\label{sec:training}
We employ the same training strategy as Goodfellow~\emph{et al.}~\cite{goodfellow2014generative} to train the generator and discriminator. We alternate between one step of optimizing discriminator and one step of optimizing generator for training our method. We did not use any more tricks in our training. We adopt Adam~\cite{kingma2014adam} with the learning rate $10^{-4}$ for both generator and discriminator. In the experiments, we observe that the proposed model is easy to train and stable during training. This may be due to the generation strategy, which can progressively generate point clouds from low resolution to high resolution. In our progressive generator, the low-resolution network is easier to train due to the simple shape with fewer points. The stability of the low-resolution network contributes to the training of the high-resolution network. Furthermore, we also found that our method convergences quickly during training. %We will release our code on Github after the paper is published.

\section{More Visualization and Quantitative Results}\label{sec:visualization}

\subsection{Generated multi-resolution point clouds}\label{sec:multi_resolutions}
\begin{figure}
	\centering
	\includegraphics[width=0.9\linewidth]{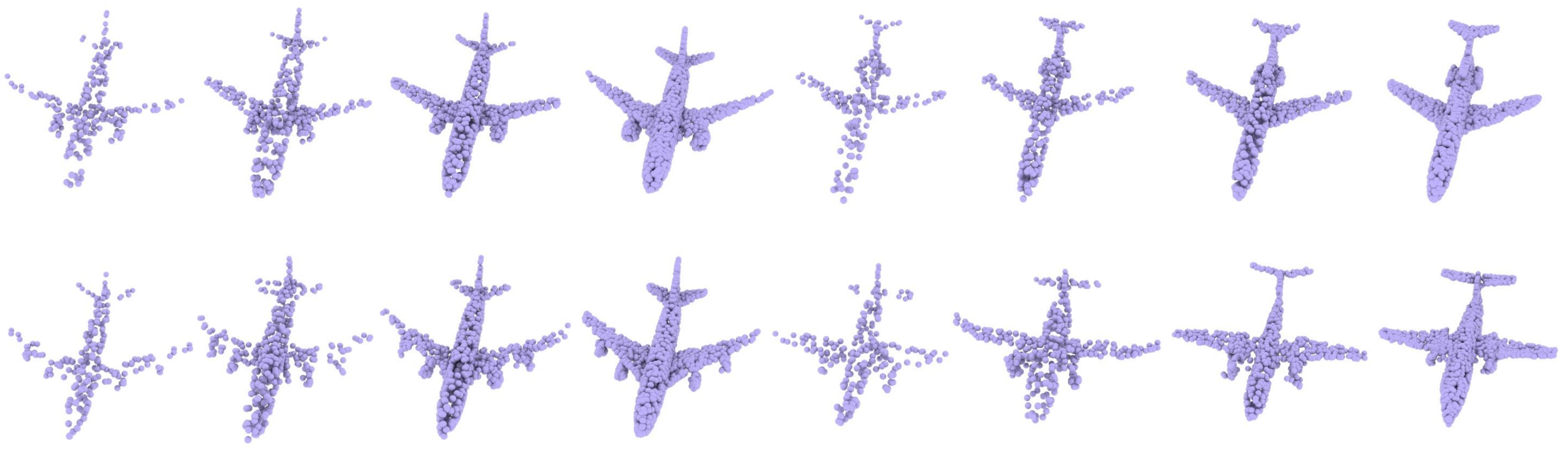}
	\caption{Visualization results of the ``Airplane'' category. The resolutions are 256, 512, 1024, and 2048, respectively.}
	\label{fig:vis_airplane}
\end{figure}

\begin{figure}
	\centering
	\includegraphics[width=0.9\linewidth]{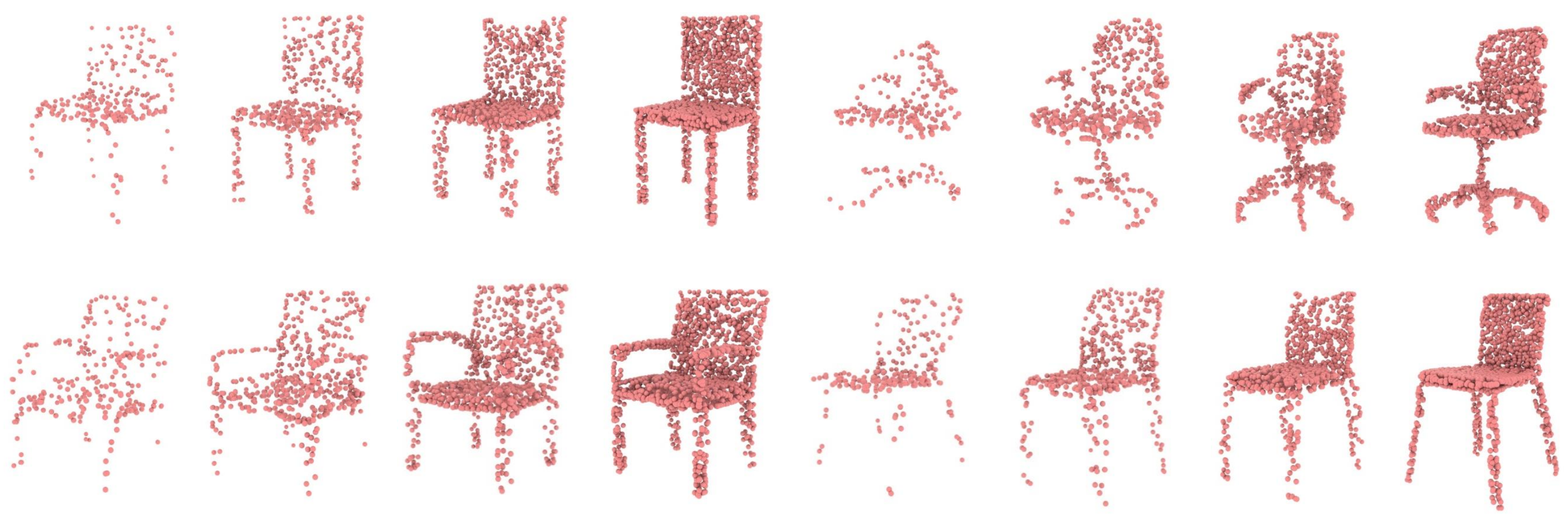}
	\caption{Visualization results of the ``Chair'' category. The resolutions are 256, 512, 1024, and 2048, respectively.}
	\label{fig:vis_chair}
\end{figure}

\begin{figure}
	\centering
	\includegraphics[width=0.9\linewidth]{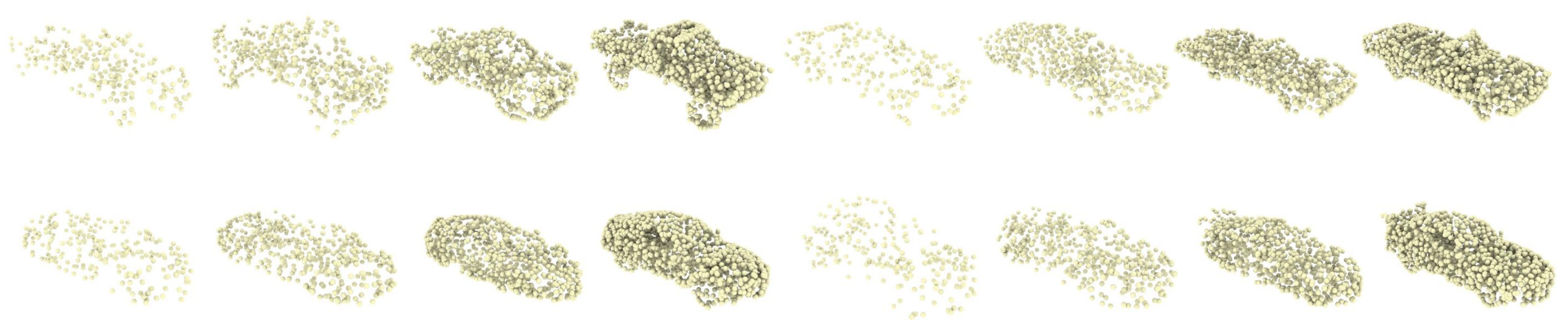}
	\caption{Visualization results of the ``Car'' category. The resolutions are 256, 512, 1024, and 2048, respectively.}
	\label{fig:vis_car}
\end{figure}

\begin{figure}
	\centering
	\includegraphics[width=0.9\linewidth]{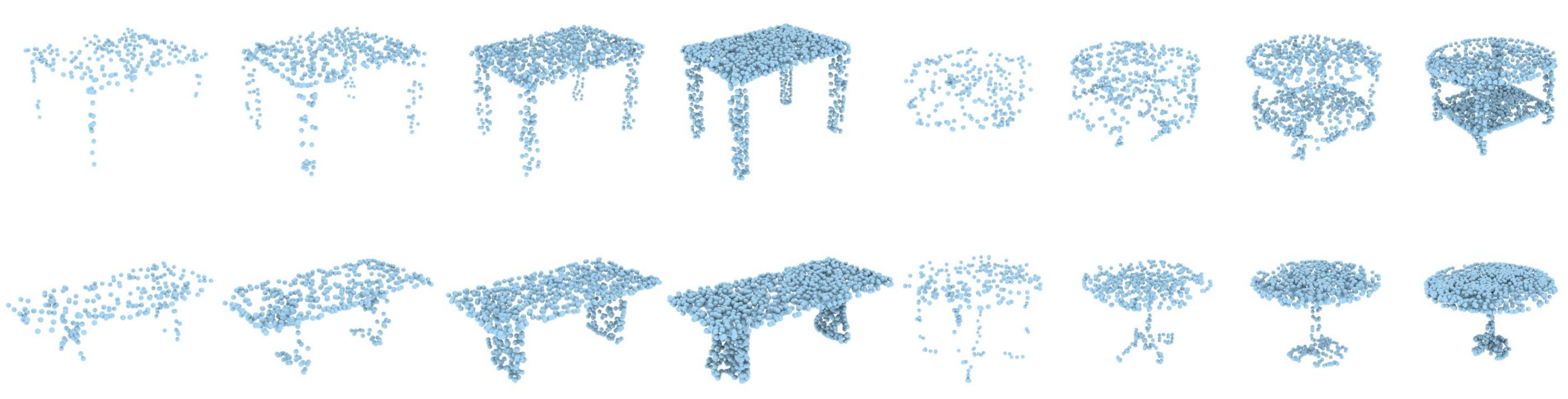}
	\caption{Visualization results of the ``Table'' category. The resolutions are 256, 512, 1024, and 2048, respectively.}
	\label{fig:vis_table}
\end{figure}

\begin{figure}
	\centering
	\includegraphics[width=0.9\linewidth]{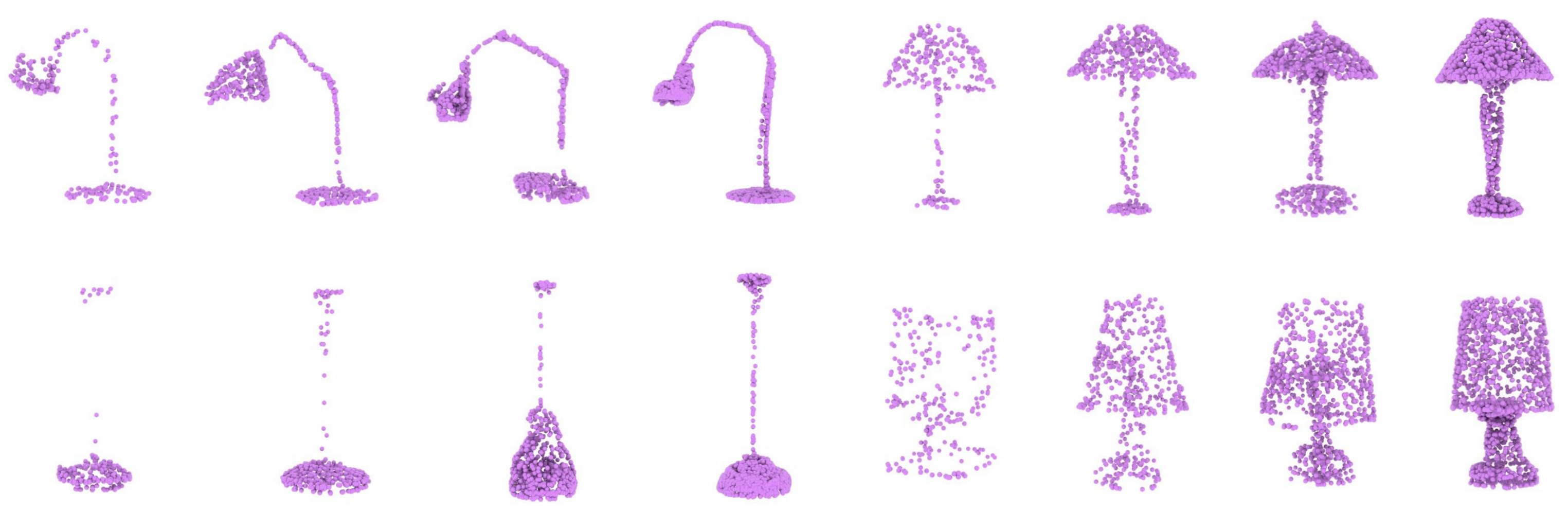}
	\caption{Visualization results of the ``Lamp'' category. The resolutions are 256, 512, 1024, and 2048, respectively.}
	\label{fig:vis_lamp}
\end{figure}

\begin{figure}
	\centering
	\includegraphics[width=0.9\linewidth]{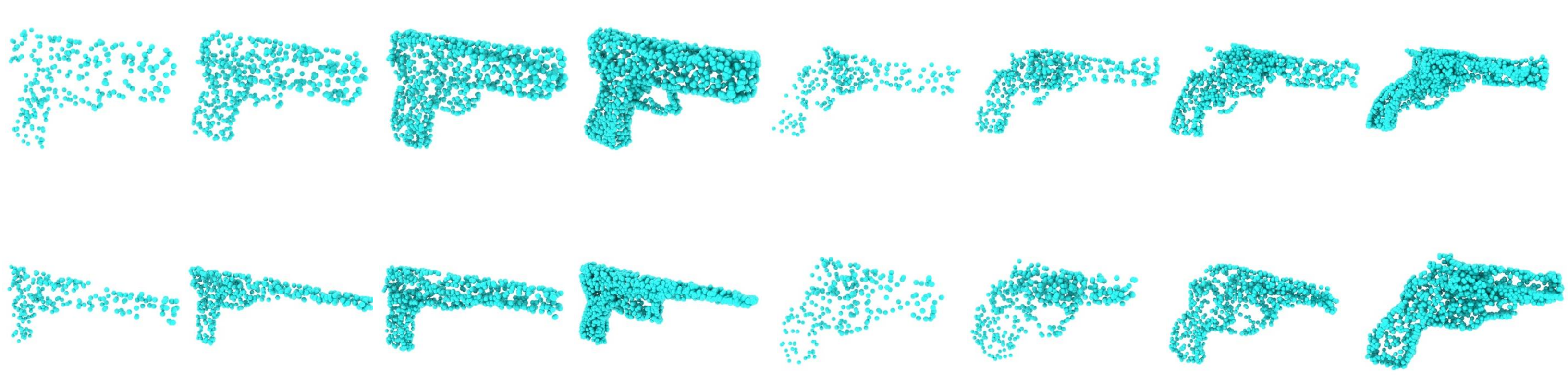}
	\caption{Visualization results of the ``Pistol'' category. The resolutions are 256, 512, 1024, and 2048, respectively.}
	\label{fig:vis_pistol}
\end{figure}

\begin{figure}
	\centering
	\includegraphics[width=0.9\linewidth]{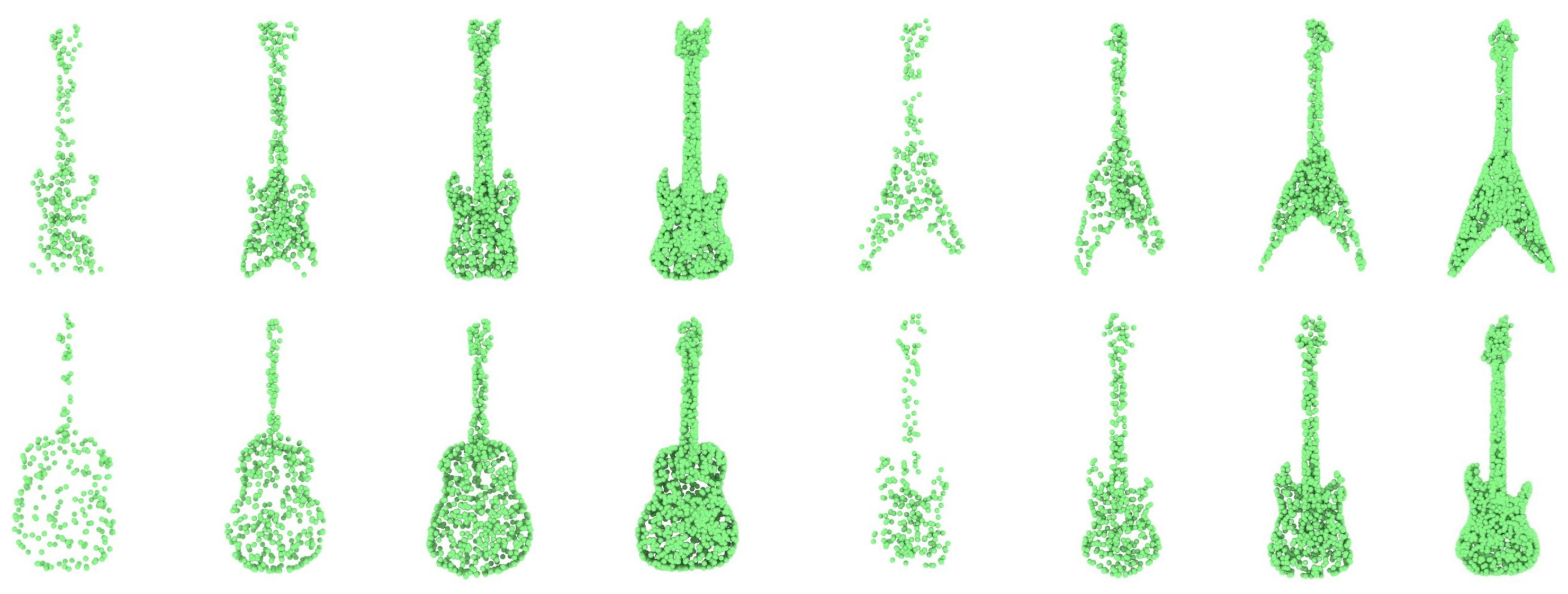}
	\caption{Visualization results of the ``Guitar'' category. The resolutions are 256, 512, 1024, and 2048, respectively.}
	\label{fig:vis_guitar}
\end{figure}

We visualize more generated point clouds of four resolutions 256, 512, 1024, and 2048, respectively. As shown in Figs.~\ref{fig:vis_airplane},~\ref{fig:vis_chair},~\ref{fig:vis_car},~\ref{fig:vis_table},~\ref{fig:vis_lamp},~\ref{fig:vis_pistol}, and~\ref{fig:vis_guitar}, they contain seven categories including ``Airplane'', ``Chair'', ``Car'', ``Table'', ``Lamp'', ``Pistol'', and ``Guitar''. From the figure, it can be clearly seen that the generated multi-resolution point clouds are consistent.

\subsection{Features in progressive deconvolution network.}\label{sec:subsec_feavis}
\begin{figure}
	\begin{center}
		\includegraphics[width=0.88\linewidth]{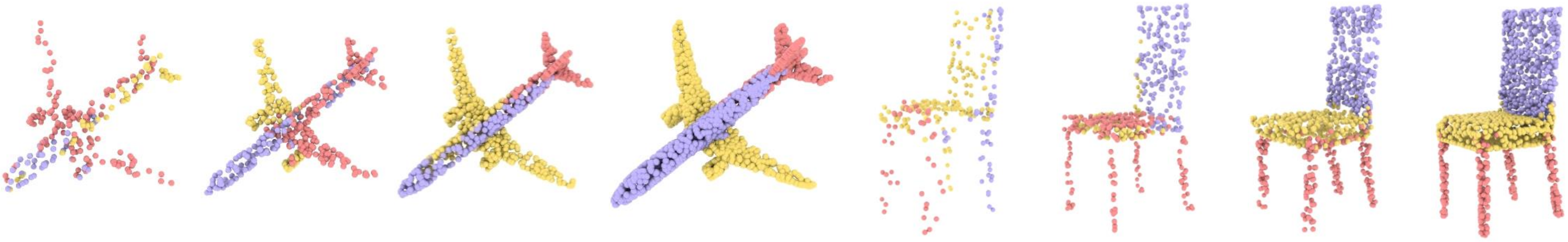}
	\end{center}
	\caption{Visualization results for features in four deconvolution networks by $k$-means clustering, colored onto the corresponding point clouds. On the ``Airplane'' and ``Chair'' categories, each category has resolutions 256, 512, 1024, and 2048, respectively. Different parts of the chairs are distinguished more clearly as the resolution of the deconvolution network increases.}
	\label{fig:deconv_fea}
\end{figure}

We analyze the outputs of different resolutions of the deconvolution network and visualize them in the feature space. As shown in Fig.~\ref{fig:deconv_fea}, we visualize the generated point clouds  on the ``Airplane'' and ``Chair'' categories with the outputs of four deconvolution networks by using $k$-means clustering in the feature space.

\subsection{Three views of the generated point clouds}\label{sec:three_views}
\begin{figure}
	\centering
	\includegraphics[width=0.9\linewidth]{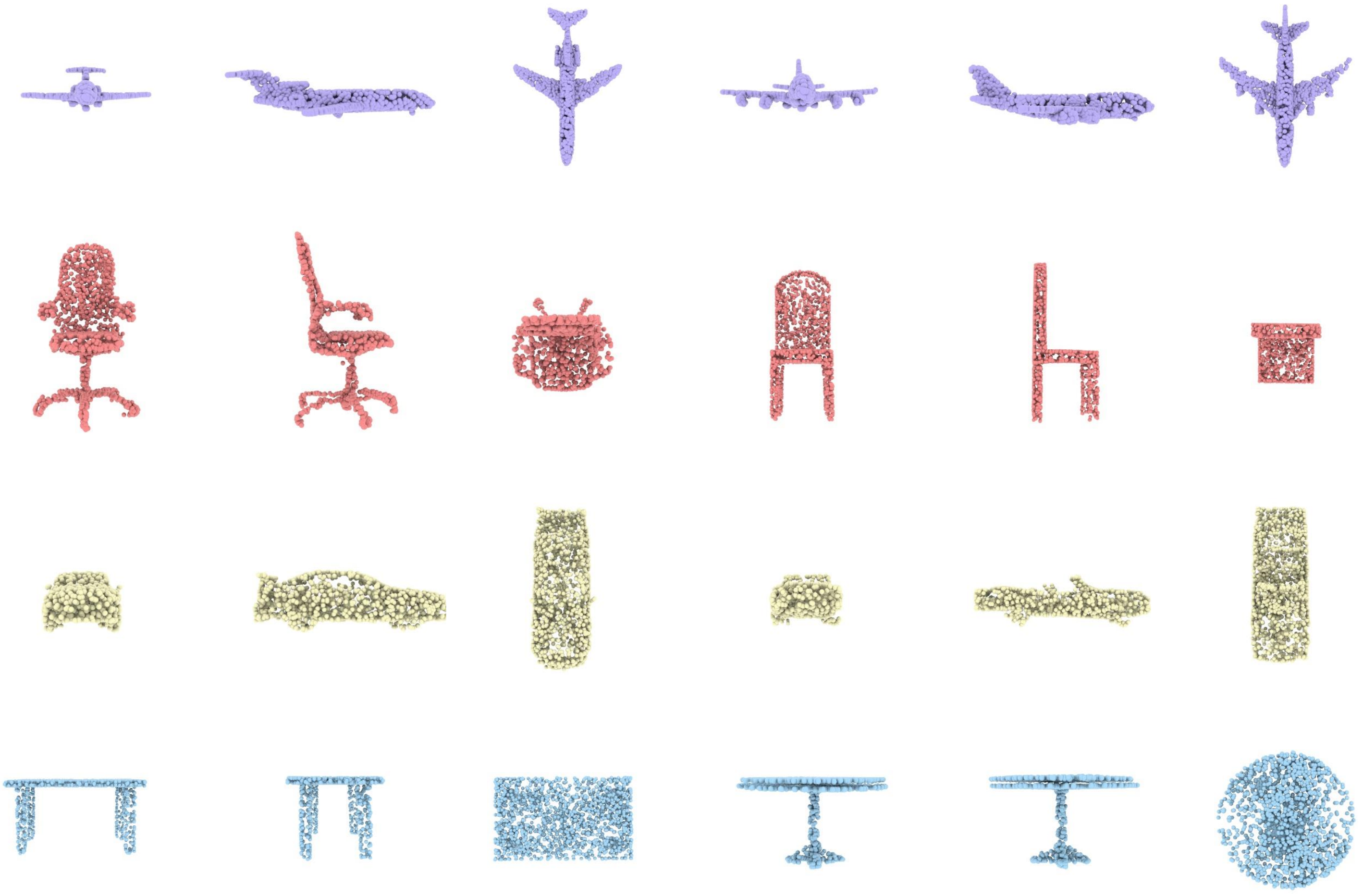}
	\includegraphics[width=0.87\linewidth]{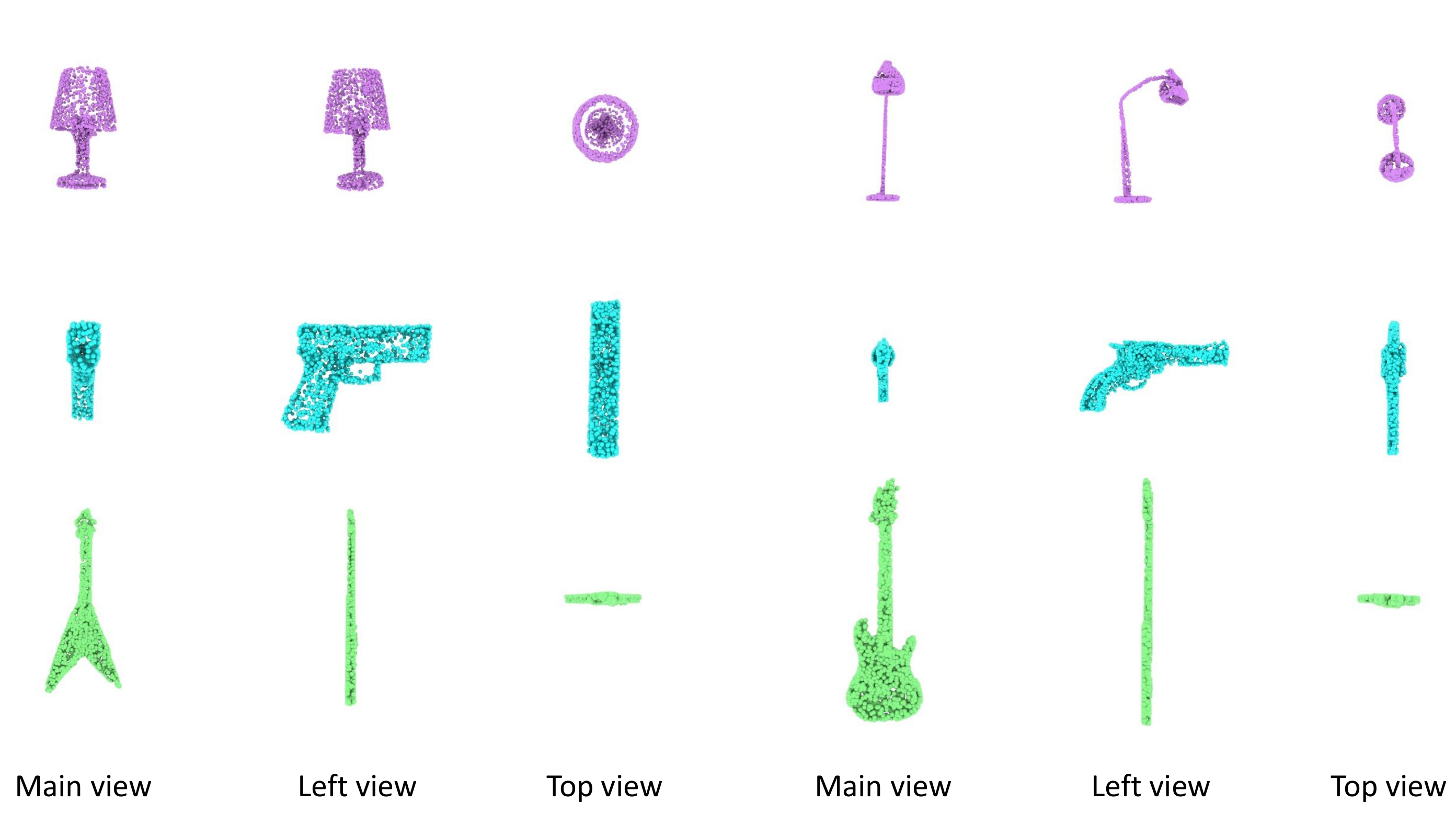}
	\caption{Three views of generated point clouds including ``Airplane'', ``Chair'', ``Car'' ``Table'', ``Lamp'', ``Pistol'', and ``Guitar'' categories. The resolution of each generated point cloud is 2048.}
	\label{fig:vis_3views}
\end{figure}

As shown in Fig.~\ref{fig:vis_3views}, we visualize seven categories including ``Airplane'', ``Chair'', ``Car'', ``Table'', ``Lamp'', ``Pistol'', and ``Guitar''. It can be seen that our generated point clouds have realistic shapes comparing to real point clouds.

\subsection{Quantitative results of the generated point clouds}\label{sec:metric_car}
As shown in Table.~\ref{tab:metric_car}, we provide the metric results on the ``Car'' category and the mean results of all 16 categories. On the ``Car'' category, metric results of our PDGN are comparable to the results of PointFlow~\cite{pointflow}. This may be because the ``Car'' category does not have many thin structures. On all 16 categories, our PDGN is better than r-GAN~\cite{achlioptas2017learning}, tree-GAN~\cite{shu20193dpc}, and PointFlow~\cite{pointflow}. Note that we released the mean results of PointFlow for all 16 categories by running the official code on 16 categories.

\begin{table}
	\caption{The comparison results of different methods on the different categories. The All (16) presents the mean results of all 16 categories. The best results are highlighted in \textbf{bold}. Note that JSD scores and MMD-EMD scores are multiplied by 10$^2$. MMD-CD scores are multiplied by 10$^3$. Lower JSD, MMD-CD, MMD-EMD, 1-NNA-CD, and 1-NNA-EMD show better performance. Higher COV-CD and COV-EMD indicate better performance.}
	\begin{center}
		\resizebox{0.8\textwidth}{!}
		{ 
			\begin{tabular}{llcccccccccc}
				\toprule
				\multicolumn{1}{c}{\multirow{2}{*}{Category}} &\multicolumn{1}{c}{\multirow{2}{*}{Model}}&  \multicolumn{1}{c}{\multirow{2}{*}{JSD ($\downarrow$)\;\;\;\;}} & \multicolumn{2}{c}{MMD ($\downarrow$)} & & \multicolumn{2}{c}{COV (\%, $\uparrow$)} & & \multicolumn{2}{c}{1-NNA (\%, $\downarrow$)} \\
				\cmidrule{4-5} \cmidrule{7-8} \cmidrule{10-11}
				&\multicolumn{1}{c}{}&&CD&EMD&&CD&EMD&&CD&EMD\\
				\midrule
				
				\multirow{6}{*}{Car}&r-GAN~\cite{achlioptas2017learning}&12.8&1.27&8.74&&15.06&9.38&&97.87&99.86 \\
				&l-GAN (CD)~\cite{achlioptas2017learning}&4.43&1.55&6.25&&38.64&18.47&&63.07&88.07\\
				&l-GAN (EMD)~\cite{achlioptas2017learning}&2.21&1.48&5.43&&39.20&39.77&&69.74&68.32\\
				&PC-GAN~\cite{Li2018PointCG}&5.85&1.12&5.83&&23.56&30.29&&92.19&90.87\\
				&PointFlow~\small\cite{pointflow}&0.87&{\bf 0.91}&{\bf 5.22}&&{\bf 44.03}&{\bf 46.59}&& 60.65& 62.36\\
				&PDGN (ours)&{\bf 0.75} &1.07&5.27&&41.17&42.86&&{\bf 57.89}&{\bf 61.53} \\
				
				\midrule
				\multirow{4}{*}{All (16) }&r-GAN~\cite{achlioptas2017learning}&17.1&2.10&15.5&&58.00&29.00&&-&- \\
				&tree-GAN~\cite{shu20193dpc} &10.5&1.80&10.7&&{\bf 66.00}&39.00&&-&- \\
				&PointFlow~\cite{pointflow} & 8.42&2.34&7.82&&45.85&52.32&&58.01&60.22 \\
				&PDGN (ours) &{\bf 6.45}&{\bf 1.68}&{\bf 6.21}&&56.58&{\bf 53.65}&&{\bf 56.85}&{\bf 59.31} \\
				%&PDGN (ours) &-&-&-&&-&-&&-&- \\t
				\bottomrule     
			\end{tabular}
		}
	\end{center}
	\label{tab:metric_car}
\end{table}

\section{Visualization Comparison with PointFlow}\label{sec:comparison}
\begin{figure}
	\centering
	\includegraphics[width=0.85\linewidth]{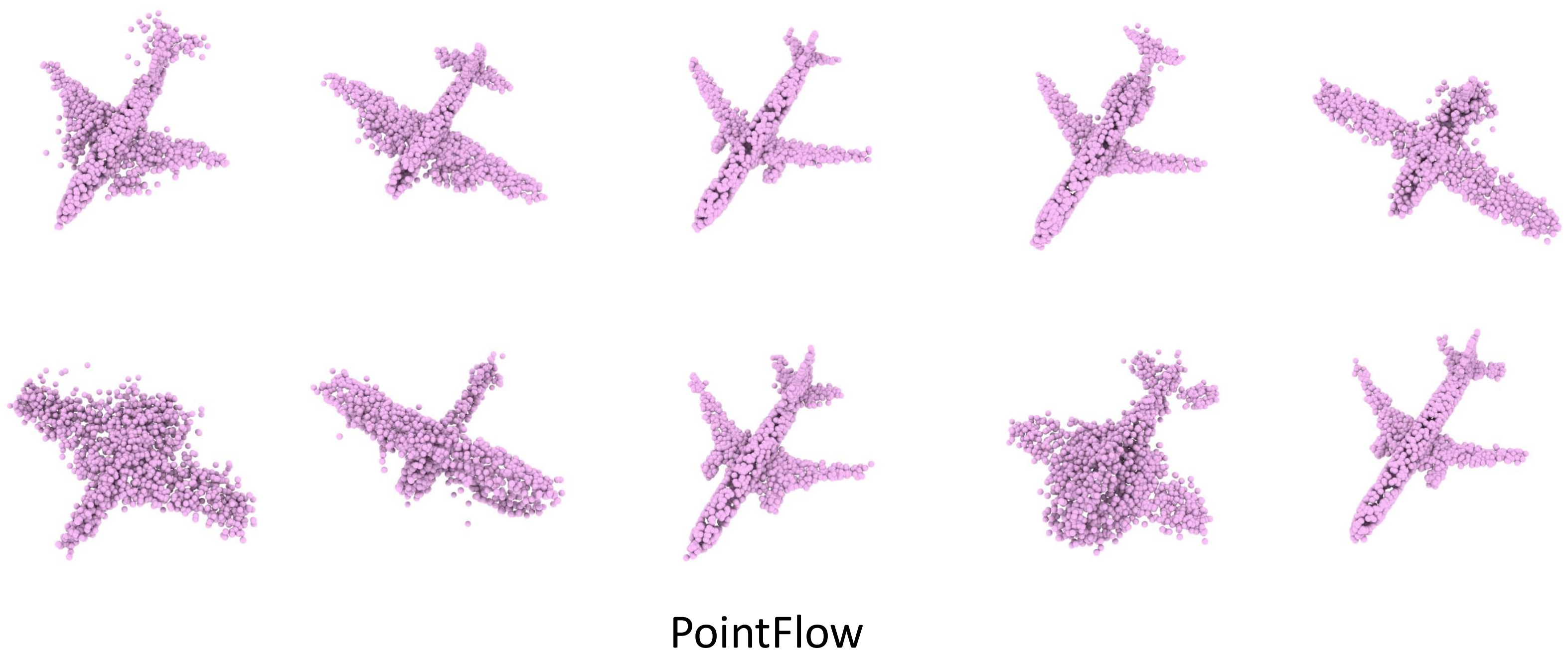}
	\includegraphics[width=0.85\linewidth]{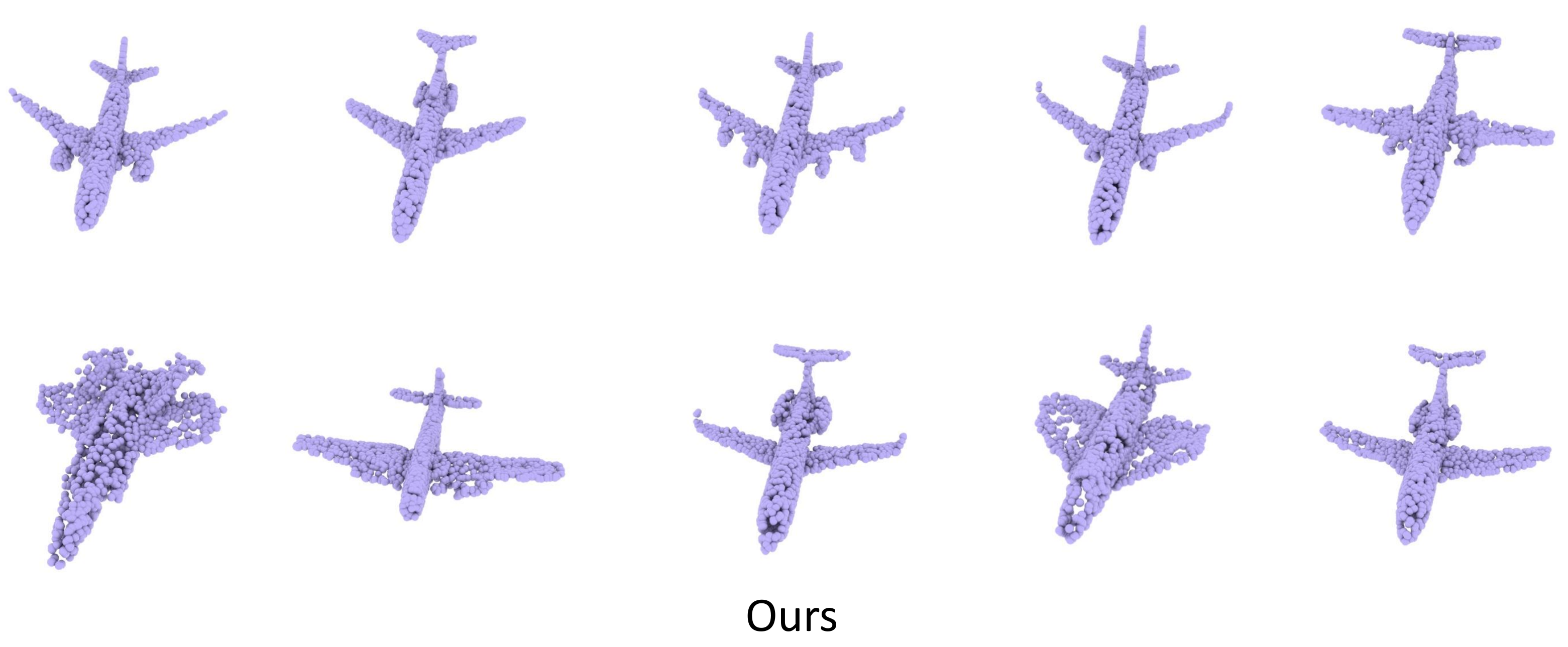}
	\caption{Visualization results of our method and PointFlow on the ``Airplane'' category. The resolution of each generated point cloud is 2048.}
	\label{fig:cmp_our_airplane}
\end{figure}

\begin{figure}
	\centering
	\includegraphics[width=0.9\linewidth]{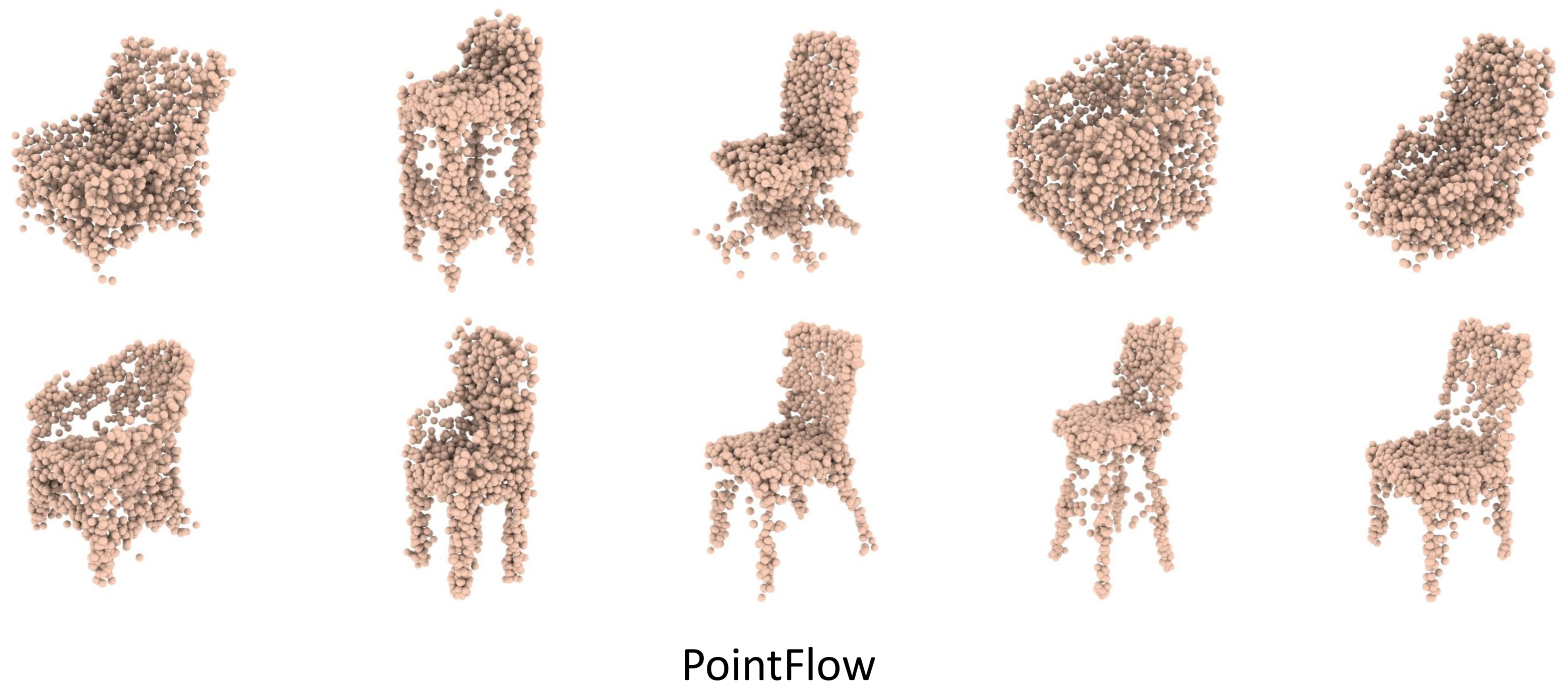}
	\includegraphics[width=0.85\linewidth]{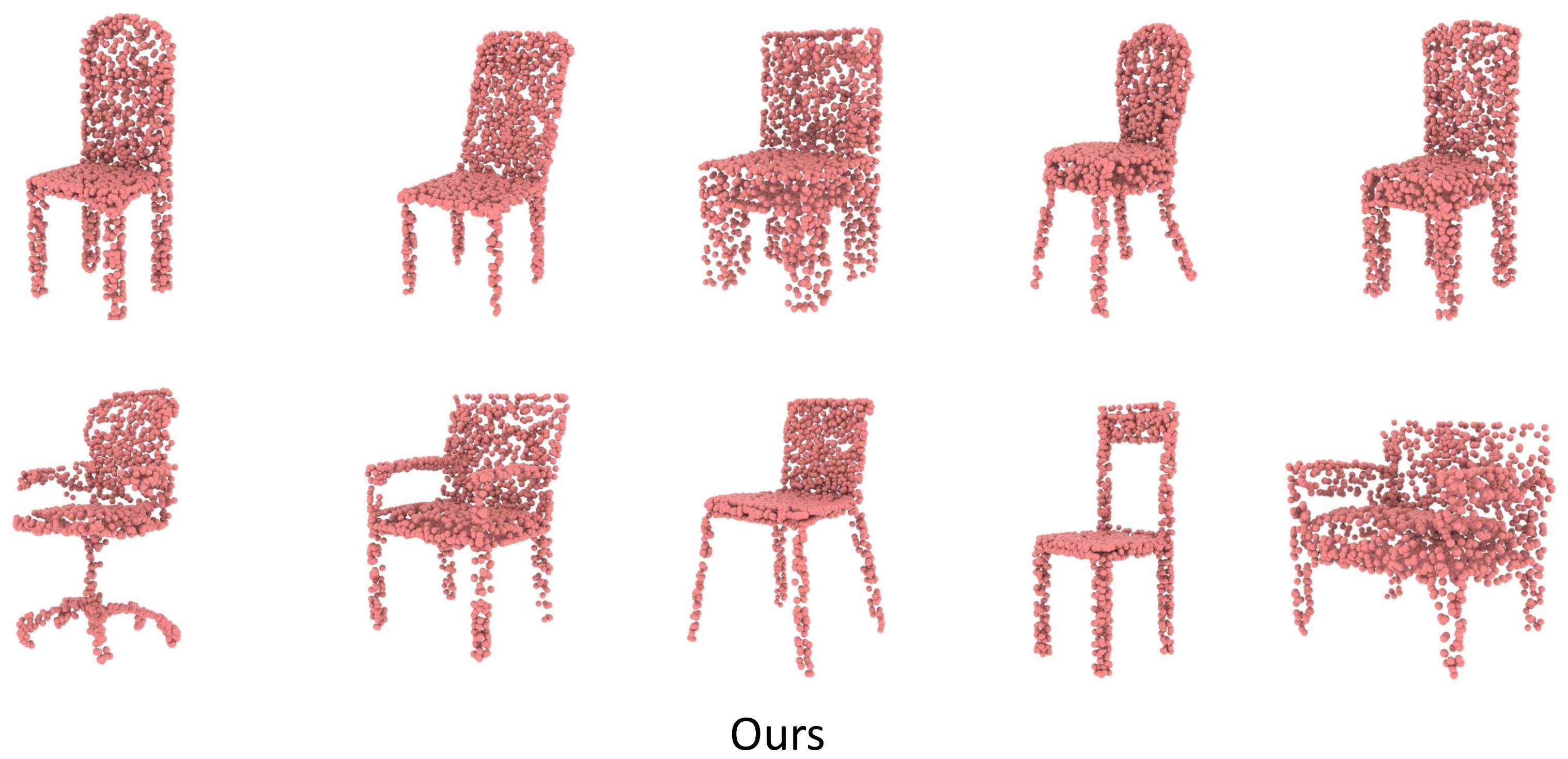}
	\caption{Visualization results of our method and PointFlow on the ``Chair'' category. The resolution of each generated point cloud is 2048.}
	\label{fig:cmp_our_chair}
\end{figure}

As shown in Figs.~\ref{fig:cmp_our_airplane} and~\ref{fig:cmp_our_chair}, we compare with the advanced method PointFlow~\cite{pointflow}. Specifically, we use the trained model provided by PointFlow on GitHub to generate point clouds. As mentioned in PointFlow~\cite{pointflow}, it fails for the cases with many thin structures (like chairs). However, due to the progressive generator from the low resolution to the high resolution, our method performs well on point clouds with thin structures (like chairs).


\end{document}